\pdfoutput=1%For ARXIV

\documentclass{article} % For LaTeX2e
\usepackage{nips14submit_e,times}
\usepackage{hyperref}
\usepackage{url}
\usepackage{amsmath}
\usepackage{subfigure}
\usepackage{graphicx}
\usepackage{tablefootnote}
\usepackage{threeparttable}
\usepackage{multirow}
\usepackage{tcolorbox}
\usepackage{algorithm}
\usepackage{algorithmic}
\usepackage{float}
\usepackage{booktabs}
\usepackage{arydshln}
\usepackage{colortbl}
\usepackage{amssymb}

\title{Every Part Matters: Integrity Verification of Scientific Figures Based on Multimodal Large Language Models}

\author{
Xiang Shi \\
School of Information Management\\
Wuhan University\\
Wuhan, Hubei 430072 \\
\texttt{coding@whu.edu.cn} \\
\And
Jiawei Liu \\
School of Information Management\\
Wuhan University\\
Wuhan, Hubei 430072 \\
\texttt{laujames2017@whu.edu.cn} \\
\AND
Yinpeng Liu \\
School of Information Management\\
Wuhan University\\
Wuhan, Hubei 430072 \\
\texttt{yinpengliu@whu.edu.cn} \\
\And
Qikai Cheng \\
School of Information Management\\
Wuhan University\\
Wuhan, Hubei 430072 \\
\texttt{chengqikai@whu.edu.cn}  \\
\And
Wei Lu \thanks{Corresponding Author.}\\
School of Information Management\\
Wuhan University\\
Wuhan, Hubei 430072 \\
\texttt{weilu@whu.edu.cn}  \\
\\
}

\nipsfinalcopy % Uncomment for camera-ready version

\begin{document}

\maketitle

\begin{abstract}
This paper tackles a key issue in the interpretation of scientific figures: the fine-grained alignment of text and figures. It advances beyond prior research that primarily dealt with straightforward, data-driven visualizations such as bar and pie charts and only offered a basic understanding of diagrams through captioning and classification. We introduce a novel task, \textbf{Figure Integrity Verification}, designed to evaluate the precision of technologies in aligning textual knowledge with visual elements in scientific figures. To support this, we develop a semi-automated method for constructing a large-scale dataset, \textbf{Figure-seg}, specifically designed for this task. Additionally, we propose an innovative framework, \textbf{E}very \textbf{P}art \textbf{M}atters (\textbf{EPM}), which leverages Multimodal Large Language Models (MLLMs) to not only incrementally improve the alignment and verification of text-figure integrity but also enhance integrity through analogical reasoning. Our comprehensive experiments show that these innovations substantially improve upon existing methods, allowing for more precise and thorough analysis of complex scientific figures. This progress not only enhances our understanding of multimodal technologies but also stimulates further research and practical applications across fields requiring the accurate interpretation of complex visual data.
\end{abstract}

\section{Introduction}

The adage 'A picture is worth a thousand words' resonates particularly in the realm of scientific communication, where the ability to elucidate complex ideas through diagrams and graphs is not just beneficial but essential. Traditionally, however, in scientific publications, figures have often served merely as supplements to textual content, with their significance frequently overshadowed by the text itself. Consequently, for a long time,  these scientific figures have not garnered as much attention from the image research community as natural images have. This perspective is now changing as the complexity of scientific data, concepts, and methodologies increases, particularly in fields like biology and medicine, where figures are vital for illustrating intricate ideas and fostering interdisciplinary collaboration. Moreover, an increasing number of researchers recognize that interpreting scientific figures is crucial for advancing powerful multimodal technologies and poses unique challenges compared to interpreting natural images \cite{kahou2017figureqa,hu2023mplug}. To significantly advance this research, this paper attempts to address a fundamental challenge in figure understanding—the fine-grained alignment of text and figures.

Text-image alignment, also formally known as visual grounding, has been a fundamental task in multimodal research. Its goal is to match natural language text with specific visual content in images. Specifically, in natural image-focused alignment tasks, the main focus is on recognizing people and objects in the images, as shown in Figure 1(a). The features of these visual entities can be effectively captured by visual encoding models such as Yolo \cite{jiang2022review} and CLIP \cite{radford2021learning}. However, aligning scientific figures with text is even more complex. Scientific figures often represent a complex integration of multiple visual elements, each linked to specific texts, terms, or concepts, as shown in the left side of Figure 1(b). Moreover, scientific figures vary widely. They range from morphological diagrams depicting biological or geological forms, to spectral line charts presenting chemical or physical properties, and system diagrams like flowcharts or framework diagrams that illustrate complex processes and relationships. Without sufficient domain knowledge, it is difficult for even humans to understand the meaning expressed by each element of a scientific figure, let alone machines or models.

In response to this challenge, current advancements primarily focus on data figures such as pie, line, and bar charts, leading to the creation of visual question-answering and captioning datasets including ChartQA \cite{masry2022chartqa}, FigureQA \cite{kahou2017figureqa}, and SciCap \cite{hsu2021scicap}. These efforts have achieved coarse-grained alignment between text and scientific figures and have inspired the design of multimodal large language models (MLLMs) such as mPLUG-PaperOwl \cite{hu2023mplug}, Vary \cite{wei2023vary}, and TextHawk \cite{yu2024texthawk}, which serve to parse these figures. Despite these advancements, existing research has two major limitations. First, the understanding of figures remains confined to data-driven charts, with little exploration into figures containing complex domain-specific information such as framework diagrams or flowcharts. Second, while studies have generally captured a generalized representation of information within figures, they still significantly lack in perceiving detailed aspects. Existing models often misinterpret these figures, either by perceiving nonexistent elements or by misidentifying the spatial and semantic features of figure components \cite{tong2024eyes}.

Recognizing the limitations of current research, this paper focuses on the following research question: How can textual knowledge in scientific publications be aligned with visual elements in complex scientific figures? Furthermore, around this issue, we approach the challenge of text-figure alignment from three dimensions: tasks, data, and models. At the task level, we propose a novel task—\textbf{Figure Integrity Verification}—to assess the model's ability to understand scientific figures. Namely, a robust figure understanding model should be able to efficiently identify each visual element in a figure, align the textual knowledge with each element, and provide textual knowledge supplementation for elements in the figure that are not aligned when no related information is found in the text, as shown in Figure 1(b). Secondly, at the data level, we design a semi-automated method to align figure elements with textual terms, resulting in the creation of a large-scale aligned dataset—\textbf{Figure-seg}, which includes descriptions of the spatial and semantic information of figure elements. Finally, at the model level, we develop a text-figure alignment framework around MLLMs, termed '\textbf{E}very \textbf{P}art \textbf{M}atters' (\textbf{EPM}). This framework requires the model to simultaneously utilize multiple capabilities, such as understanding text and figure elements, and modeling the mapping between text and figure elements, to achieve the target task. Simultaneously, we design a method based on analogical reasoning, utilizing citations from scientific publications, to supplement the description of unaligned elements in figures.

Under the guidance of the task, data, and model design methods described above, we conduct extensive experiments to validate the effectiveness of our approach. The results show that, in the context of aligning complex scientific figures, our proposed method significantly surpasses the existing state-of-the-art techniques.  Specifically, it improves text-figure alignment performance by 22.53\% on the CIoU metric and 45.13\% on the gIoU metric. It also boosts performance on detecting unaligned figure elements by 4.90\% on the CIoU metric and 4.52\% on the gIoU metric, offering superior comprehension of complex figures. Additionally, an analysis of four distinct architectures of visual understanding models reveals that incorporating background knowledge about the spatial and semantic features of elements within figures not only enhances performance significantly—yielding positive gains in approximately 70\% of the evaluated metrics—but also further highlights the unique challenges and requirements of interpreting scientific figures compared to natural images.

\begin{figure}
	\centering
		\includegraphics[scale=.6]{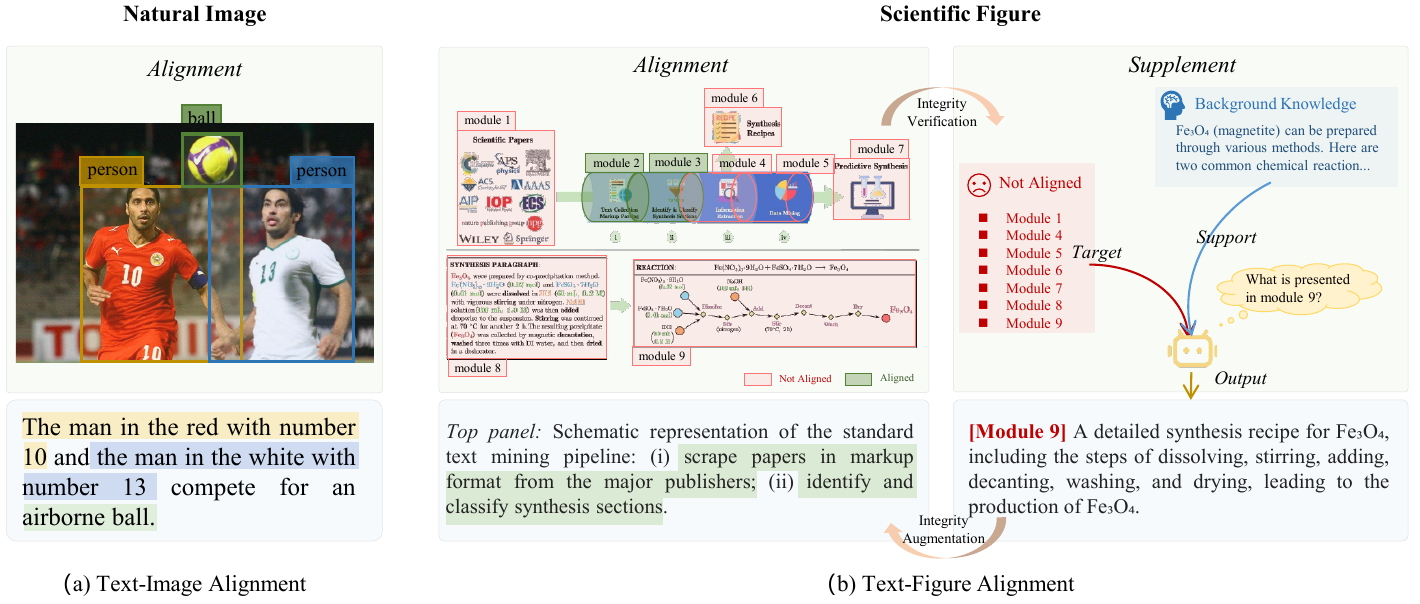}
	\caption{
Comparison of alignment tasks for natural images and scientific figures. The text-figure alignment task requires models to parse each module element within a figure, align the text, and identify unaligned elements (Integrity Verification). Additionally, it requires models to provide supplementary descriptions for unaligned elements through figure understanding (Integrity Augmentation). The natural images shown in the examples are from the CoCo dataset \cite{lin2014microsoft}, and the scientific figures are from \cite{wang2022dataset}.}
	\label{FIG:task}
\end{figure}

In summary, our contributions include:
\begin{itemize}
    \item \textbf{The Novel Task for Validating Figure Integrity.} We introduce the Figure Integrity Verification Task to enhance the capability of MLLMs in understanding complex scientific figures. This task challenges models to not only achieve precise alignment between text and figure components but also to identify figure components not mentioned in the text, thereby pushing the boundaries of current technology.
    \item \textbf{The First Fine-grained Text-Figure Alignment Dataset.} We develop a high-quality alignment dataset through a method that combines automated processes with manual verification. This dataset is crucial for performing detailed parsing, alignment, and integrity verification of complex scientific figures.
    \item \textbf{Specialized MLLM for Text-Figure Alignment.} We design a specialized MLLM to address the challenge of fine-grained alignment between textual knowledge and figure content. This model enhances the linkage between text elements and figure modules by deeply understanding and segmenting the attribute information of individual modules within figures.
    \item \textbf{A Comprehensive Framework for Figure Integrity Verification.} We propose a framework that leverages our specialized MLLM to comprehensively assess the integrity of scientific figures. We present extensive experimental results, comparing them to a series of baseline methods, to demonstrate the superiority of our approach in enhancing the comprehension and analysis of complex scientific visuals. Additionally, we design a method based on analogical reasoning to enhance the integrity of figures. \footnote{Our code and data can be found at \href{https://github.com/shixiang1a/figure_understanding}{github}.}
\end{itemize}

\section{Related Work}

The research on understanding scientific figures has increasingly gained attention from researchers in recent years. Although studies in this field are not as mature as those for natural images, the emergence of MLLMs with potent image comprehension capabilities presents new opportunities for development in this area. In this section, we provide a comprehensive review of the existing research on understanding scientific figures and MLLMs, identifying gaps in the research and the specific problems that this paper aims to address.

\subsection{Scientific figure understanding}
Although research on scientific figures commenced later than that on natural images, it has a history of over 25 years. Early explorations in this field were mostly confined to a few types of figures with relatively fixed formats, where figures and text already had clear correspondences. Researchers initially constructed formal expressions of image-text mapping manually, focusing on the structure and layout of images, as well as the natural language description methods. For instance, \cite{watanabe1998diagram} established a rule system for figure understanding based on the analysis of connections and adjacency relations of image elements in pictorial books of flora (PBF), along with the textual description methods of the images. \cite{ferguson1998telling} designed a framework named JUXTA for schematic diagrams in texts, encompassing three levels: visual, physical, and procedural. 

 As image recognition technologies such as object detection and image segmentation advanced, researchers began to explore beyond single figure types, seeking automated paths for understanding scientific figures. This led to the refinement of sub-tasks like figure syntactic parsing, figure classification, figure captioning, and figure QA. In syntactic parsing, \cite{kembhavi2016diagram} used an LSTM-based method to automatically map figures to Diagram Parse Graphs (DPG), designed specifically for figure elements and relationships. For figure classification, \cite{kembhavi2016diagram} automatically constructed a dataset named ACL-Fig, which includes 19 types of figures such as trees, graphs, and architecture diagrams, using scientific literature published by ACL. In the realm of figure captioning, \cite{hsu2021scicap,yang2023scicap+} built datasets like SciCap and SciCap+ for figure captioning research, utilizing PDF document parsing and OCR technology. \cite{singh2023figcaps} explored a reinforcement learning approach combined with domain expert feedback to generate reader-preference optimized titles. In Figure QA, researchers like \cite{kahou2017figureqa,li2023scigraphqa,zhou2023enhanced,methani2020plotqa} contributed various types of QA data for different figure types, including single and multi-turn questions, and designed figure understanding pre-training models and QA frameworks based on plot figures.

Despite the rich research in figure analysis, it remains in its nascent stages, with many studies focusing on establishing benchmarks for figure understanding and analysis tasks, lacking mature solutions. Moreover, research on scientific figures is mostly limited to plot figures used for describing experimental data results, with relatively less focus on the parsing of structural and flow diagrams that describe the content frameworks and methods of literature. To address these gaps, we specifically design a fine-grained understanding dataset and methodological framework for framework structure and flow figures, further advancing the research in scientific figure comprehension.

\subsection{Integrity Verification}
The term 'integrity verification' originates from the field of information security. Its fundamental meaning is to ensure that data or information remains unaltered from its creation, through transmission, to storage, thereby preserving its integrity \cite{barker2015profile}. Drawing on the connotations of this term, in our previous work, we introduced it into the field of information science to investigate the integrity of scientific literature \cite{shi2024integrity}. Our aim is to ensure that scientific literature, from the moment of writing, accurately and precisely conveys the intended scientific ideas to the reader. This involves ensuring that every core element within the literature, such as key terms, formulas, and figures, is clearly described and presented in a standardized manner. This paper extends this framework to explore the integrity of figure elements within the scientific literature. Although prior work in information science, particularly concerning scientific figures, has not explicitly defined the task of integrity verification, some studies have discussed the issues of scientific integrity in the creation of scientific figures \cite{bik2016prevalence,zhuang2021graphical}, which is closely related to our research.

\subsection{Multimodal large language model}
The development of LLMs has provided new perspectives for many previously challenging problems, including in the field of image understanding. Building on the foundational capabilities of text-only LLMs, researchers have developed various types of multimodal LLMs (MLLMs). One approach views LLMs as tool controllers, using natural language interaction to combine and call upon various vision processing foundation models to accomplish complex tasks in image understanding and analysis, such as with Visual ChatGPT \cite{wu2023visual} and Chameleon \cite{lu2023chameleon}. This method still relies heavily on the capabilities of traditional vision processing foundation models and does not fundamentally enhance performance in specific tasks like image understanding. Another approach involves adding image-text alignment modules to the input, embedding, MLP, and self-attention layers of the original text-only LLMs. This integration allows the features of images to merge into the LLM's textual feature space, enabling image understanding and even image generation. For instance, the PICa model \cite{yang2022empirical} uses an offline image-to-text model to convert images into descriptions, combining this with knowledge from databases to design prompts that utilize GPT-3’s understanding capabilities for Visual Question Answering (VQA). The LLava model \cite{liu2023visual} maps image features encoded by image encoders directly into LLama's \cite{touvron2023llama} text features through multimodal mappers. LLama-Adapter \cite{zhang2023llama} and LLama-Adapter2 \cite{gao2023llama} introduce learnable parameters into LLama’s transformer layers to encode image features. Meanwhile, models like BLIP2 \cite{li2023blip}, InstructBLIP \cite{dai2305instructblip}, and Qwen-VL \cite{bai2023qwen} incorporate a Q-former module to align images and text through shared self-attention layers. Furthermore, several studies have expanded upon the well-established foundation of multimodal LLMs, adapting them for various specific image understanding tasks. Examples include the LISA model \cite{lai2023lisa}, designed for image segmentation, and the mPLUG-PaperOWL model \cite{hu2023mplug}, developed for the interpretation of charts.

Inspired by research in the foundational architecture of MLLMs and their applications in image understanding tasks, we design a MLLM for interpreting scientific figures, with a focus on structure and process diagrams, from multiple dimensions including data and model architecture. This model aims to achieve pixel-level mapping between textual descriptions and figure modules, thus serving the task of integrity verification for scientific figures.

\section{Problem Formulation}
\textbf{Figure Integrity Verification.} We operate under the assumption that every figure presented in academic literature is meticulously designed to illustrate research methods, processes, or results. Each element within a figure serves a distinct purpose or conveys a specific message intended by the author. Thus, to fully communicate the ideas embodied in the figure, the function or meaning of each component should be clearly explained in the accompanying text. Motivated by this premise, our study aims to enhance the understanding of complex diagrams, such as flowcharts and framework diagrams, by meticulously aligning text with corresponding figure elements. Our approach addresses three main challenges:
\begin{itemize}
    \item \textit{Text-figure Alignment:} How to achieve fine-grained alignment between text and figure modules.
    \item \textit{Integrity Verification:} How to identify and highlight content within figures that has not been adequately described in the text.
    \item \textit{Integrity Augmentation:} How to supplement the descriptions of unaligned modules within figures through figure understanding.
\end{itemize}

\textbf{Formal expression.} Assuming a paper comprises $l$ figures $G=\{g_1,g_2,\ldots,g_l\}$ and $k$ text segments $T=\{p_1,p_2,\ldots,p_k\}$ (where $p$ can represent a term or a sentence), each figure, $g_i$, contains n independent modules $g_i=\{m_{i1},m_{i2},\ldots,m_{in}\}$. The process of solving the first problem involves establishing connections between each module, $m_{ij}$, within figures and text paragraphs, $p_j$, forming $\left(m_{ij},p_j\right)$ pairs, wherein the relationship between $m_{ij}$ and $p_j$ is many-to-many. Upon successfully linking figure modules with text, we obtain a set of modules explained by the corresponding text, $g_i^+=\{m_{ij}\in g_i|\ if\ \ (m_{ij},p_j)\ \ exists\ \ in\ \ T\}$. Then, the solution to the second problem transforms into performing set operations $g_i-g_i^+$ to derive the set of figure modules, $g_i^-$, that have not been appropriately described. Finally, the method designed to address the third problem aims to find an appropriate description for any $m_{ij}^- \in g_i^-$.

\section{Method}
In this section, we present a detailed exposition of how to endow a generic MLLM with the capability to comprehend and validate the integrity of scientific figures. This is accomplished by commencing with the construction of data (Section \ref{4.1}) and subsequently progressing to the design of the model architecture (Section \ref{4.2}).

\begin{figure}[h!]
	\centering
		\includegraphics[scale=.5]{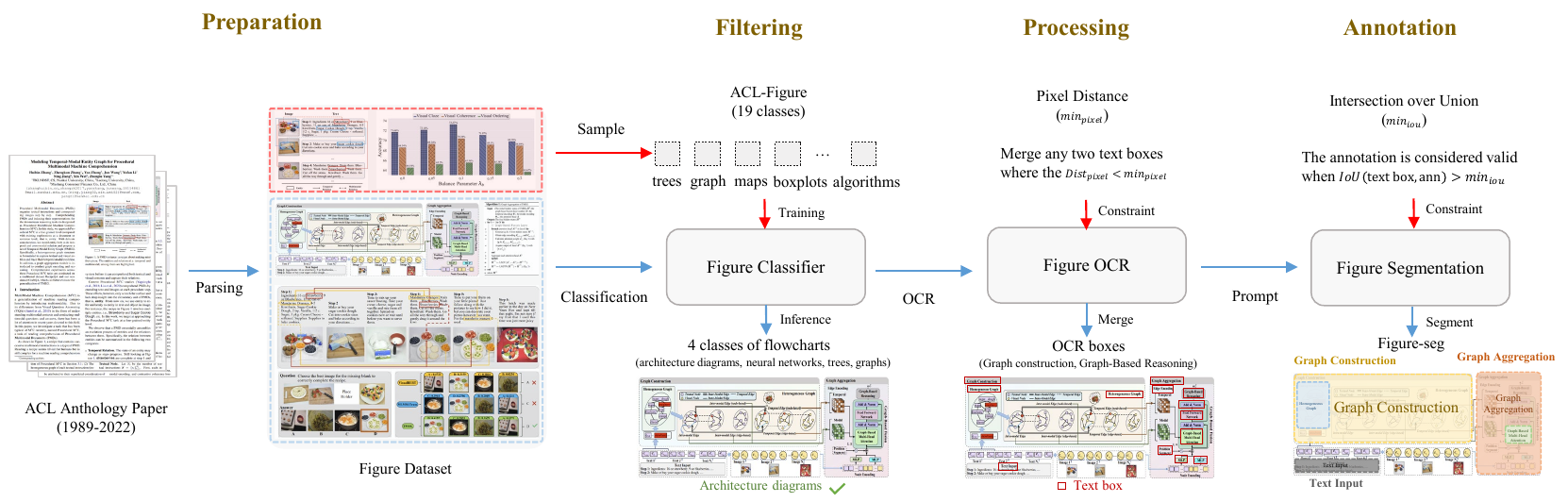}
	\caption{Overview of the construction process for a fine-grained dataset aligning scientific figures with text, comprising four sequential stages: preparation, filtering, processing, and annotation. The collaboration between automated tools and human annotators facilitates the execution of the filtering, processing, and annotation stages.}
	\label{FIG:dataset}
\end{figure}

\subsection{Dataset construction}
\label{4.1}
In constructing our dataset, we delineate the process into two distinct stages: \textit{figure segmentation} and \textit{semantic enhancement}.The former aims to achieve fine-grained alignment between text and specific modules within figures, whereas the latter builds upon this foundation by introducing spatial and semantic information for precise localization of modules that appear multiple times within figures, thereby enhancing the accuracy of the dataset.
\subsubsection{Figure segmentation}
The process of figure segmentation is illustrated in Figure \ref{FIG:dataset}, encompassing four stages: \textit{preparation}, \textit{filtering}, \textit{processing}, and \textit{annotation}. The specific operations for each stage are as follows and the setting of parameters refers to Appendix \ref{section:E}.

\textbf{Preparation Phase.} Our dataset is constructed from a rich collection of open-access papers provided by the Association for Computational Linguistics (ACL), which are abundant in model frameworks and flowcharts. In this phase, we meticulously parse the original PDF documents to extract the figures embedded within. To streamline this process, we utilize the highly efficient tool, PDFFigures 2.0 \cite{clark2016pdffigures}, renowned for its ability to proficiently extract content from scientific literature. Notably, this tool possesses the unique capability to distinguish between figures and tables, offering an essential preliminary step in filtering figures relevant for our further analysis.

\textbf{Filtering Phase.} We refine a specialized classifier, CLIP-fig, to discern between different types of figures, leveraging the ACL-fig dataset \cite{karishma2023acl}, derived from ACL papers, as its training foundation. This classifier adeptly categorizes scientific figures into 19 distinct categories, demonstrating a commendable accuracy rate of 81.49\% when evaluated against test datasets. In our research, we concentrate on five primary categories: algorithms, architecture diagrams, neural networks, trees, and graphs, which encapsulate the majority of flowchart representations pertinent to this field. Figures that do not fit into these specified categories are systematically excluded, ensuring that only the relevant figures are retained for in-depth parsing.

\textbf{Processing Phase.} During this stage, we utilize the PaddleOCR tool \cite{li2022pp} to extract key terms and sentence fragments from the target figures, which act as anchor texts for pinpointing modules within the figures. Recognizing the inherent limitations of OCR tools, which might interpret text spanning multiple lines as discrete segments, we implement an empirical strategy to amalgamate text based on the minimum pixel distance between text boxes. This involves calculating the centroid $(x_i, y_i)$ of each text box identified by the OCR, using the coordinates of its detection box's corners. We then determine the pixel distance between centroids of different text boxes with the formula $\textit{Dist}_{pixel}=\sqrt{(x_i-x_j)^2+(y_i-y_j)^2} \ i \neq j$. If the distance $\textit{Dist}_{pixel}$ falls below a predefined threshold $min_{pixel}$, we consider these text segments as parts of the same original sentence or term. Consequently, we merge these texts and their respective detection boxes to enhance the accuracy of our figure annotation process.

\textbf{Annotation Phase.} In the annotation phase, the extracted text is fed into the FastSAM model \cite{zhao2023fast} as prompt words to achieve fine-grained localization of modules within the figures. Considering the complex representation and semantic depth of scientific figures, which differ significantly from natural images, FastSAM, although proficient with natural imagery, may face challenges in accurately segmenting modules in scientific figures. To enhance data quality and reduce the impact of potential inaccuracies, we adopt a strategy that combines empirical constraints with manual verification. Initially, we employ the centroid coordinates of modules identified from the merged text boxes as prompts for a secondary application of FastSAM's image segmentation, ensuring the segmented modules align with the text both spatially and contextually. We then calculate the Intersection Over Union (IOU) between the text boxes and segmented modules, requiring it to surpass a specified threshold, $min_{iou}$. This step verifies that the modules segmented by FastSAM based on varying prompts exhibit consistency and substantial overlap with the text boxes, suggesting reasonable accuracy. A subsequent manual review process pairs the text with its corresponding figure modules to create \textit{(text, module)} pairs. Additionally, we extract paragraphs describing the figures from their citations within the papers, simultaneously identifying and labeling figure components not mentioned in these descriptions. This process results in the creation of \textit{(text, missed module)} pairs, serving as a measure for assessing the integrity verification effectiveness of figures.

\begin{figure}[h!]
	\centering
		\includegraphics[scale=.57]{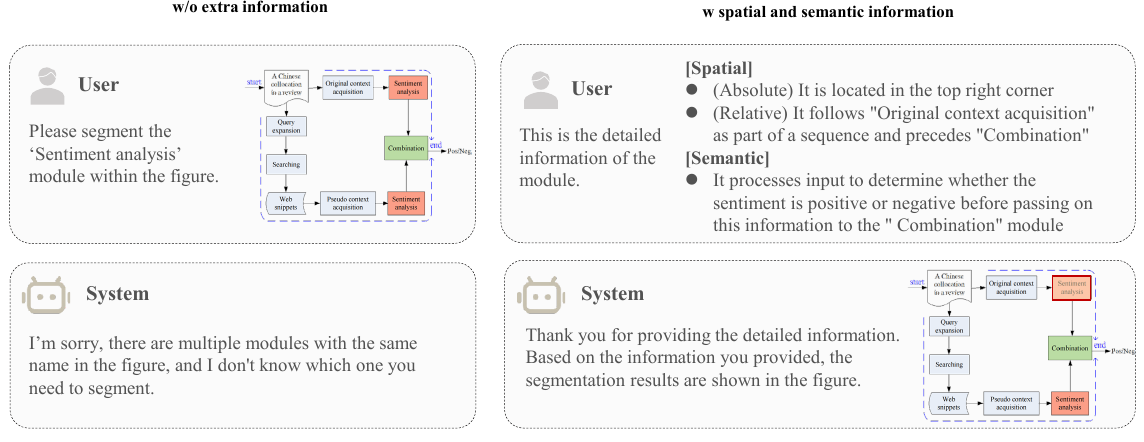}
	\caption{Instances of semantic enhancement based on the figure from the study by \cite{zhao2012collocation}. In their figure, 'Sentiment analysis' modules appear multiple times. To differentiate these modules more precisely, we incorporate spatial and semantic information.}
	\label{FIG:enhancement}
\end{figure}

\subsubsection{Semantic enhancement}
We randomly sample 1,000 scientific figures from the previously collected dataset for analysis and find that approximately 31.5\% of the figures contain multiple instances of modules with identical names. This indicates that simply inputting the name text into the semantic segment model can lead to confusion, preventing accurate identification of the corresponding modules in the figures. To address this issue, we enhance the semantic information of the text by incorporating spatial and semantic information into the constructed text-figure module alignment data. As illustrated in Figure \ref{FIG:enhancement}, spatial information includes both absolute and relative positioning, utilized respectively to describe the module's location within the entire figure and its spatial relationships with nearby modules. Relative positioning encompasses directional relationships, connectivity, and containment. Semantic information, on the other hand, delineates the role of a module within the entire figure or in relation to other relevant modules.

\begin{figure}[h!]
	\centering
		\includegraphics[scale=.7]{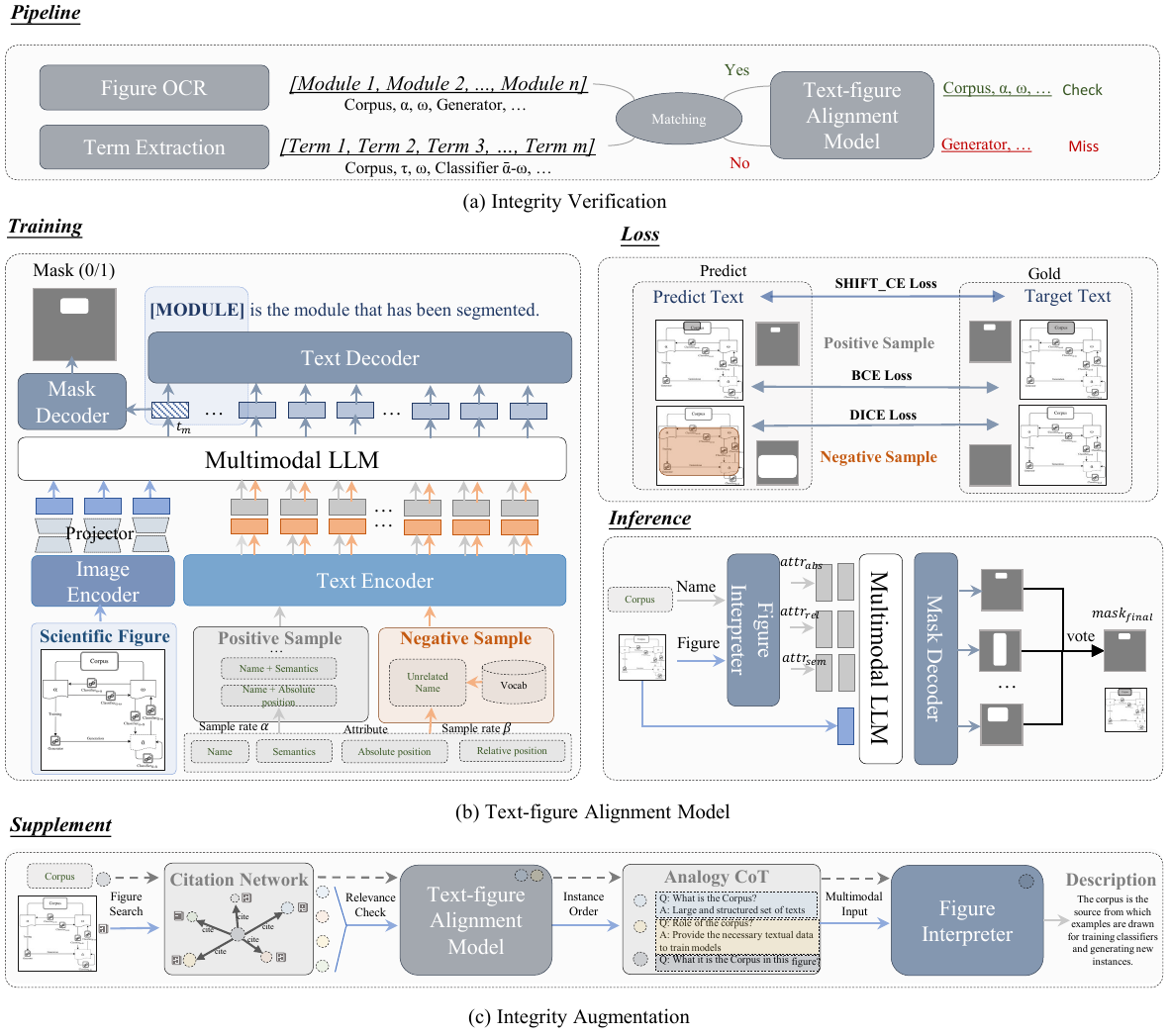}
	\caption{The overall framework for implementing the integrity verification task. Figure (a) depicts the key modules and the implementation process required to complete the task. Figure (b) elaborately illustrates the training and inference process of the core module - the Text-figure alignment Model, which is crucial for aligning figure modules with text.}
	\label{FIG:framework}
\end{figure}

\subsection{Model Architecture}
\label{4.2}
In the architectural design of the model, we draw inspiration from the LISA \cite{lai2023lisa}, basing our framework on MLLMs with capabilities in image understanding and segmentation. Specifically tailored for the unique characteristics of flowcharts and framework diagrams, we design a proprietary feature extraction module. The model is introduced in the following sections, as illustrated in Figure \ref{FIG:framework}.

\subsubsection{Input}
Assuming a scientific figure $g_i \in G$ and an accompanying paragraph $p_j \in T$ delineate module $m_{ij}$ within $g_i$. In the initial phase, leveraging a meticulously crafted instruction template alongside sampling strategies, the paragraph $p_i$ from the academic paper undergoes transformation into a question input $q_{ij}$, thereby rendering it more intelligible to the model. This input $q_{ij}$ encompasses a distinctive symbol <image>, signifying image input, coupled with question text that encapsulates details regarding the attributes of module $m_{ij}$. Within the ambit of the textual component, diverse combinations of spatial and semantic attributes are sampled at a rate of $\alpha$, thus furnishing the model with the aptitude to assimilate assorted attribute information and ensuring the coherence of outcomes generated throughout its training trajectory. Simultaneously, the model undergoes enhancement through negative sampling predicated on the attribute information of $m_{ij}$, imbuing it with the proficiency to identify modules absent from the figures. Precisely, contingent upon the name of $m_{ij}$, names not manifested within figure $g_i$ are arbitrarily culled from a lexicon (constituted by the terminology of all modules within the dataset) in a ratio of $\beta$ as negative instances $q_{ij}^-$, deemed extraneous to the content of the figure. Conclusively, $q_{ij}$ and $q_{ij}^-$ coalesce to engender a question set $q_{ij}^*$, serving as the model's input.

\subsubsection{Encoder}

\textbf{Text encoder.} The text encoder employs the tokenizer pre-trained with the Vicuna model \cite{vicuna2023} to process the input text $q_{ij}^*$ by segmenting it into a sequence of tokens. Subsequently, these tokens are mapped into the textual space through an embedding matrix $F_{txt_enc}\in \mathbb{R}^{\left|v\right|\times d_e}$ to form the text embedding $e_{q_{ij}}$.
\begin{equation}
    e_{q_{ij}^*} = \mathcal{F}_{txt\_enc}(q_{ij}^*)
\end{equation}
The notation $|v|$ denotes the number of tokens in the vocabulary used by the tokenizer, $d_e$ represents the dimensionality of the token vectors configured, and $\mathcal{F}_{txt\_enc}$ signifies the text embedding function.

\textbf{Image encoder.} The image encoder is adeptly crafted to project images of various resolutions into vector representations with precision, facilitating an advanced understanding and extraction of image features. This encoder combines a visual encoding model, $\mathcal{F}_{vis\_enc}$, with a multimodal mapping layer, $\textit{MLP}_{mm}$. $\mathcal{F}_{vis\_enc}$ maps images to a vector space, enhancing feature extraction, while the multimodal mapping layer bridges the gap between image and text spaces. By mapping visual information to the textual domain, the encoder allows images to be interpreted by language models similarly to how text is processed. This integration significantly boosts the model's ability to comprehend and analyze visual content effectively.
\begin{equation}
    e_{g_i} = \textit{MLP}_{mm}(\mathcal{F}_{vis\_enc}(g_i))
\end{equation}

Following the encoding phase, the 
 text $e_{q_{ij}^*}$ and image vectors $e_{g_i}$ are concatenated to form a comprehensive input vector $e_{in_{ij}}$, which is then fed into multiple transformer layers for the integrated understanding of text and image.
\begin{equation}
    e_{in_{ij}} = \textit{Concat}([e_{g_i};e_{q_{ij}^*}])
\end{equation}

\subsubsection{Decoder}
\textbf{Text decoder.} The text decoder decodes features, $e_{out_{ij}}$, learned through multiple transformer layers into output text, $a_{ij}$, via an autoregressive mechanism. In our scenario, the structure of the model's output text is predefined as $a_{ij}=\{\mathrm{task\ description\ }t_r+\mathrm{\ attribute\ description\ }t_a+\mathrm{\ special\ token\ }t_m\}$. The task description refers to the part of the text that describes the task to be completed by the model in response to the composite question, $q_{ij}^\ast$. There are primarily two types of tasks: one is the text-figure alignment task, and the other is the attribute recognition task for figure modules. The attribute description refers to the spatial and semantic information of the module $m_{ij}$ in the figure. This description is used exclusively in the attribute recognition task. The special token, denoted as [MODULE], is employed to represent the module mask predicted by the model. This token will be used exclusively in the text-figure alignment task.
\begin{equation}
    a_{ij} = \arg\max P(t_r + t_a + t_m | q_{ij}^*)
\end{equation}

\textbf{Mask decoder.} The mask decoder, comprising multiple Transformer layers and mapping layers, primarily functions to decode the feature representation of the special token [MODULE] in the output text $a_{ij}$ into a binary image mask encoding, $mask_{pred_{ij}}$ (0/1). To accomplish this task, it is first necessary to index the feature $e_{mask_{ij}}$ corresponding to [MODULE] from the output features $e_{out_{ij}}$, followed by the prediction of the mask.
\begin{equation}
    e_{mask_{ij}} = \mathcal{F}_{idx}(e_{out_{ij}}, t_m)
\end{equation}
\begin{equation}
    mask_{ij} = \mathcal{F}_{mask\_dec}(e_{mask_{ij}})
\end{equation}

\subsubsection{Training}

\textbf{Loss.}
During the training phase, the loss is composed of text generation loss ($\mathcal{L}_{txt}$) and mask generation loss ($\mathcal{L}_{mask}$). The $\mathcal{L}_{txt}$ employs the commonly used shift cross-entropy loss (Shift\_CE) in generative language models, with a weight coefficient $\lambda_1$ assigned to modulate the model's focus on generating text and image masks. $a_{pred_{ij}}$ and $a_{gold_{ij}}$ respectively denote the predicted text output by the model and the target text predefined in the dataset.
\begin{equation}
\mathcal{L}_{txt} = \lambda_1\textbf{Shift\_CE}(a_{pred_{ij}}, a_{gold_{ij}}) 
\end{equation}
$\mathcal{L}_{mask}$ is constituted by a combination of binary cross-entropy (BCE) and DICE loss, addressing the alignment of the predicted masks $mask_{pred_{ij}}$ with the gold masks $ mask_{gold_{ij}}$ from both global and microscopic perspectives. These components of loss are respectively regulated by weight coefficients $\lambda_2$ and $\lambda_3$ to adjust their influence on the model's learning process.
\begin{equation}
\mathcal{L}_{mask} = \lambda_2\textbf{BCE}(mask_{pred_{ij}}, mask_{gold_{ij}}) + \lambda_3\textbf{DICE}(mask_{pred_{ij}}, mask_{gold_{ij}})
\end{equation}
Finally, $\mathcal{L}_{txt}$ and $\mathcal{L}_{mask}$ are combined to constitute the total loss $\mathcal{L}$, which is utilized for fine-tuning the parameters of the model.
\begin{equation}
\mathcal{L} = \mathcal{L}_{txt} + \mathcal{L}_{mask}
\end{equation}

\textbf{Setting.} We employ the pre-trained LLaVA-1.5 model as our base model, CLIP-ViT-Large as the image embedding model, and the SAM mask decoder as the image decoding model. During the training phase, we freeze the parameters of the base model, the image embedding model and the multimodal mapping layer. LoRA parameters are embedded alongside the linear mapping layers for each query and key within the base model, in conjunction with the mask decoder, to fine-tune the model. The detailed parameter setting in the training phase can be found in Appendix \ref{section:F}.
\subsubsection{Inference}

\begin{figure}[ht!]
    \begin{minipage}{0.65\textwidth}
        \begin{algorithm}[H]
    \caption{Chain-of-Attribute (CoA)}
    \begin{algorithmic}[1]
     \STATE \textbf{Input:} $D^{+}$: Positive examples from segmentation data, $G$: Figure, $M$: Module name, $\mathcal{I}$: Figure interpreter trained on $D^{+}$
    \STATE \textbf{Output:} $\textit{Mask}_{final}$: Final Segmentation

    \STATE $\textit{Exist} \gets \mathcal{M}(G, M)$ \ \textcolor{gray}{// Determine if $M$ exists in $G$}
    % \STATE \ \ \ \textcolor{gray}{// Determine if $M$ exists in $G$ using model $\mathcal{M}$}
    \IF{$\textit{Exist} = \text{False}$}
    \STATE $\textit{Mask}_{final} \gets \emptyset$
    \ELSE
        \STATE $(attr_{abs}, attr_{rel}, attr_{sem}) \gets \mathcal{I}(G, M)$
        \STATE \ \ \ \textcolor{gray}{// Output module $M$'s attributes using interpreter $\mathcal{I}$}
        \STATE $\textit{Mask}_{abs} \gets \mathcal{M}(G, \textit{Concat}(M, attr_{abs}))$ 
        \STATE $\textit{Mask}_{rel} \gets \mathcal{M}(G, \textit{Concat}(M, attr_{rel}))$ 
        \STATE $\textit{Mask}_{sem} \gets \mathcal{M}(G, \textit{Concat}(M, attr_{sem}))$  
        \STATE \ \ \ \textcolor{gray}{// Obtain segmentations by feeding attributes to $\mathcal{M}$}
        \STATE $\textit{Mask}_{final} \gets \text{Vote}(\textit{Mask}_{abs}, \textit{Mask}_{rel}, \textit{Mask}_{sem})$ 
        \STATE \ \ \ \textcolor{gray}{// Synthesize final mask by voting among $\textit{Masks}$}
    \ENDIF
    \end{algorithmic}
\end{algorithm}
    \end{minipage}
    \hfill
    \subfigure[Flowchart of the CoA Process]{
    \begin{minipage}{0.3\textwidth}
        \centering
        
        \includegraphics[width=\textwidth]{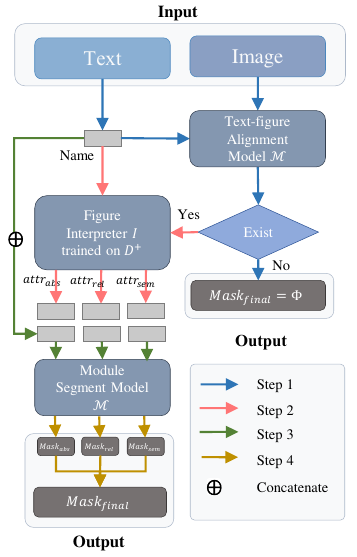} % Replace 'example-image-a' with your image file name
        
    \end{minipage}}
    \caption{Detailed explanation of the Chain-of-Attribute (CoA) reasoning process.}
    \label{FIG:coa}
\end{figure}

\textbf{Chain-of-Attribute (CoA).} Due to the limitations in the level of detail provided by paper paragraphs describing modules within images, the model lacks sufficient information input during the inference process, akin to what is available during training, leading to a decrease in inference accuracy. To address this issue, we guide the model to enrich its input through thought chains. Unlike simply adding "Let's think step by step" to encourage the model to construct thought chains on its own, we explicitly design thought chains based on attribute recognition (Chain-of-Attribute). This involves leveraging the model's image comprehension capabilities to refine descriptions of spatial and semantic features, thereby enhancing inference accuracy, as shown in Figure \ref{FIG:coa}.

CoA can be regarded as a specialized version of CoT-SC \cite{wang2022self}. CoT-SC achieves model inference through the construction and voting of multiple chains of thought, whereas CoA accomplishes inference by identifying, combining, and voting on the spatial and semantic attributes of modules within a figure. LLMs take $q_{ij}^*$ as input and predict the intermediate values $attr_{ij} \sim P(attr_{ij}|q_{ij}^*)$, then concatenate $q_{ij}^*$ with $attr_{ij}$ to form $q_{ij}^*$, which is subsequently input into the model to predict the mask of the figure module segment. Finally, the prediction outcome is formed through a majority vote based on the results predicted from different attribute information (absolute position $attr_{abs}$, relative position $attr_{rel}$ and semantic $attr_{sem}$).

\textbf{Integrity verification pipeline.} Based on the construction of the text-figure alignment model, we design a pipeline strategy to achieve the objective of integrity verification, as depicted in \ref{FIG:framework}(a). In this pipeline, the model's capabilities in three aspects are required: image understanding, text understanding, and multimodal understanding. First, image understanding evaluates the model's ability to segment anything or perform OCR on the modules within the figure, providing the foundation for subsequent alignment and integrity verification. Second, text understanding assesses the model's ability to recognize key scientific terms or concepts in the text, allowing them to be matched with elements in the figure. Finally, multimodal understanding involves finely aligning the parsed image with the text, resulting in the final integrity verification outcome. In this study, to achieve optimal results in the final integrity verification, we employ multiple models to collaboratively accomplish the task.

\textbf{Integrity augmentation.} The method of providing descriptions for unaligned modules is shown in \ref{FIG:framework}(c). Our approach is based on a simple idea: just as the creation of a scientific publication is often inspired by the publications it cites, the construction of figures might follow a similar pattern. Therefore, we can use figures and texts from previous research, along with our understanding of the figure itself, to provide explanations for any unaligned module within a figure. Based on this idea, our integrity augmentation process includes four steps: 
\begin{itemize}
    \item \textbf{Figure Retrieval.} Extract corresponding scientific figures and texts based on the citations in the publication.
    \item \textbf{Relevance Check.} Use a text-figure alignment model to identify figures relevant to the target module.
    \item \textbf{Multimodal Input Construction.} Organize the relevant figures and their descriptive texts in a QA format to form the contextual input. In this part, the questions related to the descriptive texts of the corresponding figures are generated from the extracted texts using a text understanding model. For the target module in the current figure, the questions are posed by an MLLM acting in the role of a reader from various perspectives.
    \item \textbf{Analogical Reasoning.} Input the analogical text-figure pairs into the figure interpreter to answer the dimensional information of the module, and finally summarize to form the final description.

\end{itemize}

\section{Experiment}
In this section, we present a comprehensive evaluation of our research, structured as follows: first, we describe the experimental setup; next, we detail the results of our model tests; we then discuss the findings from ablation studies to assess the robustness and essential components of our model. Additionally, we explore the method's transferability across different disciplines and provide examples of its application in analyzing scientific figures from various fields.

\subsection{Experimental setup}

\subsubsection{Dataset}
\textbf{Dataset Configuration.} For the training and evaluation of our model, we utilize a training set, a validation set, and two distinct types of test sets, containing 13,761, 1,000, 1,000, and 100 entries respectively. The first test set, Test MS, assesses the model's ability to align text with figure modules. The second test set, Test IV, focuses on the model's proficiency in detecting figure modules not described in the text. Test IV is particularly challenging, featuring nearly 1,000 modules that require identification, including a significant number of undescribed modules. This setup tests the model's comprehension and segmentation capabilities at a higher standard. Detailed statistics of these datasets are presented in Table \ref{Tbl:stastic}.

\begin{table}[h!]
\centering
\caption{Detailed statistical information of the data. Herein, 'Num.' denotes the total count, 'Avg.' signifies the average excluding the number of entries, 'Checked Num.' and 'Missed Num.' respectively represent the quantities of modules with corresponding descriptions and those without.}\label{Tbl:stastic}
\begin{tabular}{cccccc}
\hline
Object & Metric &Train & Validation & Test MS & Test IV \\
\hline
Entries & Num. & 13,761 & 1,000 & 1,000 & 100\\
\hline
Images & Num.(Avg.) & 9,770 (0.710) & 969 (0.969) & 980 (0.980) & 100 (1.000)\\
\hline
\multirow{3}*{Modules} & Num.(Avg.)  & 13,786 (1.001) & 1,002 (1.002) & 2,003 (2.003) & 978 (0.978)\\
&Checked Num.& 13,786 & 1,002 & 1,003 & 278\\
&Missed Num. & - & - & 1,000 & 700\\
\hline
\end{tabular}
\end{table}

\subsubsection{Baseline}
To ensure a fair and rigorous comparison, we select seven state-of-the-art (SOTA) models from the last two years as our baselines. These models are recognized for their performance in tasks like semantic segmentation, referring segmentation, reasoning segmentation, segment anything, and object detection: For semantic segmentation, \textbf{OVSeg} \cite{liang2023open} is a powerful open-vocabulary generalist model that maps key terms in text to modules in images without being constrained by specific categories. For referring segmentation, \textbf{GRES} \cite{liu2023gres} can segment multiple targets by integrating image regions and text features while filtering irrelevant text. \textbf{X-Decoder} \cite{zou2023generalized} and \textbf{SEEM} \cite{zou2024segment} achieve multimodal understanding and referring segmentation through learnable query features and interactions among text, image, and visual features. For reasoning segmentation, \textbf{LISA} \cite{lai2023lisa} is based on the "Embedding as Mask" concept, merging fixed mask token outputs from the image understanding model with a mask decoder to execute content segmentation in images. For segment anything, \textbf{FastSAM} \cite{zhao2023fast}, a faster iteration of the SAM model \cite{kirillov2023segment}, provides a fiftyfold speed improvement and combines comprehensive segmentation with prompt-guided selection. For object detection, \textbf{Qwen-VL} \cite{bai2023qwen} is a multimodal model for image understanding and object detection, using a visual-language adapter trained on tasks like Captioning, VQA, Grounding, and OCR.

\subsubsection{Metric}

We employ two sets of evaluation metrics to assess the performance of the models. 

\textbf{Image Segmentation Metrics.} The first set comprises metrics for evaluating the model's image segmentation capabilities, such as the cumulative intersection over the cumulative union (cIoU) and the average of all per-image Intersection-over-Unions (gIoU), following the work of \cite{lai2023lisa}. Higher values of these metrics indicate that the model is more accurate in segmenting image modules based on textual information. These metrics are utilized in both text-figure alignment and integrity verification tasks to evaluate the model's ability to locate modules described by text and those not described by text, respectively. Assuming we have $N$ images, each image contains a predicted region and a true region. For the $i^{th}$ image, let $A_i$ denote the area of the predicted region, $B_i$ represent the area of the true region, and $I_i$ signify the area of the intersection between the predicted and true regions. The calculation methodologies for cIoU and gIoU are as follows:

\begin{equation}
    cIoU=\frac{\sum_{i=1}^{N}I_i}{\sum_{i=1}^{N}\left(A_i+B_i-I_i\right)}
\end{equation}
\begin{equation}
    gIoU=\frac{1}{N}\sum_{i=1}^{N}\left(\frac{I_i}{A_i+B_i-I_i}\right)
\end{equation}

\textbf{Detection Metrics.} Precision (P), Recall (R), and F1-score (F1) are used exclusively for integrity verification tasks, assessing the detection capabilities for modules not described by text. Assuming we have N figures, each figure contains several predicted unaligned modules and true unaligned modules. For the $i^{th}$ figure, let $TP_i$ denote the number of true positives (correctly predicted unaligned modules),  $FP_i$ represent the number of false positives (incorrectly predicted unaligned modules), and $FN_i$ signify the number of false negatives (missed true unaligned modules). The calculation methodologies for Precision (P), Recall (R), and F1-score (F1) are as follows:

\begin{equation}
    Precision\left(P\right)=\frac{\sum_{i=1}^{N}{TP}_i}{\sum_{i=1}^{N}{{(TP}_i+{FP}_i)}}
\end{equation}

\begin{equation}
    Recall\left(R\right)=\frac{\sum_{i=1}^{N}{TP}_i}{\sum_{i=1}^{N}{{(TP}_i+{FN}_i)}}
\end{equation}

\begin{equation}
    \mathrm{F1-score}\left(F1\right)=2\times\frac{P\times R}{P+R}
\end{equation}

\subsection{Main result}
As detailed in Table \ref{tbl1}, our evaluation covers the performance of each model in text-figure alignment and integrity verification tasks. Our method shows a notable improvement in text-figure alignment, achieving increases of 22.53\% in cIoU and 45.13\% in gIoU metrics over FastSAM. This superior performance is primarily due to our method's enhanced ability to detect figure-irrelevant information. Despite these gains, it is important to note that our model's recognition of figure-related modules does not yet match FastSAM. This discrepancy can be partially attributed to the use of negative samples in our training, which might compromise the model’s ability to recognize positive samples. Additionally, the dataset construction was influenced by FastSAM’s capabilities, potentially biasing its performance. In tasks testing integrity verification, our method surpasses existing models, with improvements ranging from 4\% to 9\% over FastSAM. Notably, the performance in segmenting all modules within figures is higher than that in identifying only undescribed modules, with cIoU and gIoU metrics better by 4.90\% and 4.52\%, respectively. This indicates that undescribed modules present a more significant challenge, a trend consistent across various baseline models. When comparing results between the two tasks, it becomes clear that current models excel at aligning simple figure modules. However, their performance declines when faced with multiple identical modules within the same figure, suggesting a limitation in handling complex scenarios.

\begin{table}[h!]
\caption{Comparison of baseline and our models in text-figure alignment and integrity verification, detailing abilities in segmenting all modules (Every.) and detecting undescribed modules (Miss.). In text-figure alignment, parentheses denote proficiency in segmenting positive samples, while in integrity verification, they represent the model's maximum ability to detect undescribed modules. $\ddag$  
 denotes the model fine-tuned on the Figure-seg dataset.}\label{tbl1}
 \scalebox{0.98}{
\begin{tabular}{ccccccc}
\hline
Model & \multicolumn{2}{c}{Text-figure Alignment} & \multicolumn{4}{c}{Integrity Verification}\\
\cline{2-7}
& \multicolumn{2}{c}{Image} & \multicolumn{2}{c}{Image (Every.)}& \multicolumn{2}{c}{Image (Miss.)}\\
\cline{2-7}
 & cIoU $\uparrow$ & gIoU $\uparrow$ &  cIoU $\uparrow$ & gIoU $\uparrow$ & cIoU $\uparrow$ & gIoU $\uparrow$ \\
\hline
OVSeg &  3.48 (6.81) & 5.74 (11.60)  & 5.07 & 7.10 & 5.71 (4.56) & 3.94 (6.02) \\
\rowcolor{gray!20} 
$\text{OVSeg}^\ddag$ &  10.55 (17.77) & 13.58 (24.26)  & 11.58 & 15.19 & 10.13 (10.87) & 10.98 (13.73) \\
Qwen-VL &  2.39 (5.11)   & 11.37 (3.92) & 4.89 & 3.14 & 4.98 (4.49) & 1.93 (3.04) \\
\rowcolor{gray!20} 
$\text{Qwen-VL}^\ddag$ &  11.04 (14.74)   & 8.07 (15.04) & 9.96 & 10.86 & 4.77 (9.42) & 1.91 (10.02) \\
GRES &    1.14 (5.22) & 3.48 (3.06) & 2.30 & 4.88 & 4.98 (5.15) & 1.20 (2.43) \\
\rowcolor{gray!20} 
$\text{GRES}^\ddag$ &  2.15 (3.03)   & 33.65 (1.19) & 3.04 & 1.07 & 2.26 (2.84) & 0.35 (0.90) \\
X-Decoder   & 4.34 (6.28)& 30.51 (3.72) & 3.57  & 2.04 & 3.25 (3.98) & 1.15 (1.90) \\
\rowcolor{gray!20} 
$\text{X-Decoder}^\ddag$ & 2.45 (4.90)    & 5.71 (2.86)  & 4.01 &4.48  & 4.44 (3.67) &  2.54 (3.61) \\
SEEM &     3.97 (7.32) & 12.23 (10.05) & 6.24 & 5.80 & 4.97 (5.40) & 3.03 (4.66) \\
\rowcolor{gray!20} 
$\text{SEEM}^\ddag$ &  2.85 (5.23)   & 18.47 (3.84) & 4.37 & 3.00 & 5.86 (3.98) & 3.08 (2.53) \\
LISA &  26.25 (36.72) & 39.87 (48.24) & 21.25 & 25.20 & 19.06 (19.61) & 15.93 (22.10)\\
\rowcolor{gray!20} 
$\text{LISA}^\ddag$ &   33.50 (56.03)  & 37.08 (70.77) & 21.49 & 32.45 & 17.37 (16.90) &  19.95 (26.66)\\
FastSAM &  51.27 \textbf{(77.02)}  & 44.87 \textbf{(89.44)} & 22.28 & 33.64 & 17.63 (19.88) & 22.22 (29.80)\\
\hline
EPM & \textbf{73.80} (75.74) & \textbf{90.14} (82.18)  & \textbf{31.85} & \textbf{40.55}  & \textbf{23.02 (26.95)} & \textbf{26.81 (36.03)} \\
\hline
\end{tabular}}
\end{table}

\subsection{Ablation study}
To better analyze the impact of various components within our method on the overall performance of the model, we undertake ablation experiments by sequentially replacing the model's backbone, figure understanding, text understanding, and input modules. The outcomes of these experiments are depicted in Table \ref{tbl2} and \ref{tbl3}.

\begin{table}[h!]
\centering
\caption{Results of the ablation study indicate performance variations of the model upon substituting the base model, attribute recognition model, and combinations of attribute information injection. The baseline model for comparison is the EPM model with parameters $\alpha$ and $\beta$ set to 1, documented in the last row of the table.}\label{tbl2}
\scalebox{0.83}{
\begin{tabular}{cccccc}
\hline
Type &Model & \multicolumn{2}{c}{Text-figure Alignment} & \multicolumn{2}{c}{Integrity Verification}\\
\cline{3-6}
& & cIoU $\uparrow$ & gIoU $\uparrow$ & cIoU $\uparrow$ & gIoU $\uparrow$\\
\hline
\multirow{3}*{Base Model} &LLaVA-1.5-7b &  68.98 (70.00)  & 88.73 (78.46)  & 23.04 (\textbf{28.96}) & \textbf{24.68} (33.53)\\
&LLaVA-1.6-7b & 66.94 (67.78)  & 87.48 (76.37) & 23.04 (27.07) & \textbf{24.68} (31.67)\\
&LLaVA-1.6-13b & \textbf{71.03 (72.08)} & \textbf{89.92 (81.14)} & \textbf{25.44} (27.56) & 23.84 \textbf{(35.28)}\\
\hline
\multirow{3}*{Attribute Recognition}& LLaVA-1.5 &  63.98 (65.42)  & 86.92 (74.94) & \textbf{25.44} (26.70) & \textbf{23.84} (30.68) \\
&GPT-4 & \textbf{68.53 (69.77)}  & \textbf{88.55 (78.19)} & 24.17 \textbf{(28.48)} & 23.78 (\textbf{32.22}) \\
&mPLUG-PaperOwl & 57.00 (57.97)  & 84.28 (69.65)& 22.39 (21.64) & 20.63 (25.93) \\
\hline
\multirow{7}*{Attribute Combination}&name only &  63.78 (64.91) & 86.92 (74.93) & 19.61 (22.93) & 22.24 (29.72)\\
&name + rel pos. &  65.40 (66.59)  & 87.59 (76.27) & 23.97 (26.26) & 26.23 (30.87)\\
&name + abs pos. &  66.50 (67.45) & 87.39 (75.85) & 23.32 (24.44) & 25.97 (30.38)\\
&name + sem.   & \textbf{67.01 (68.45)}  & \textbf{87.82 (76.75)} &  24.01 \textbf{(26.89)} & \textbf{26.81 (31.44)}\\
&name + abs pos. + rel pos. & 64.91 (66.08) & 87.10 (75.31) & \textbf{25.23} (25.98) & 26.55 (30.60) \\
&name + abs pos. + sem. &  66.90 (67.65) & 87.51 (76.11) & 24.36 (25.04) & 26.76 (31.03) \\
&name + rel pos. + sem. &  66.13 (66.30) & 87.72 (76.54) & 23.41 (25.98) & 26.53 (30.60) \\
\hline

% \rowcolor{blue!10}
\multicolumn{2}{c}{EPM (w/o vote)} &65.84 (66.48) &  87.33 (75.75)&24.17 (28.48) & 23.78 (32.22)\\
\multicolumn{2}{c}{EPM (w vote)} &67.37 (68.41) & 87.74 (76.48)& 24.67 (26.33) & 23.88 (30.71)\\
\bottomrule
\end{tabular}}
\end{table}

\subsubsection{Base model performance}
To assess the impact of base model configurations on text-figure alignment and integrity verification tasks, we examine various versions and parameter settings of the LLaVA model \cite{liu2023improvedllava,liu2024llava} (details in the first row of Table \ref{tbl2}). Our findings show that the LLaVA-1.6-13B model, which represents the most advanced version, delivers superior performance across all evaluated tasks. In the text-figure alignment task specifically, LLaVA-1.6-13B demonstrates a significant enhancement, outperforming the LLaVA-1.5 model by approximately 3.7\%. However, its advantage in the integrity verification task is more modest, with an improvement of only 1.6\% over LLaVA-1.5. This disparity underscores the complexity of the integrity verification task, which not only requires the model to recognize all modules within a figure (global information) but also to precisely align text with specific figure modules (local details), testing the model's comprehensive understanding of figure content.

\subsubsection{Figure interpreter performance}
\label{section:D}
To explore the impact of the figure interpreter on overall performance, we compare the changes when using LLaVA-1.5, mPLUG-PaperOWI, and GPT-4 as interpreters for the spatial and semantic attributes of figure modules, as shown in the second row of Table \ref{tbl2}. In the text-figure alignment tasks, the models exhibit heightened sensitivity to the accuracy of attribute information. GPT-4, with its superior figure comprehension capabilities, shows notable improvements in our image understanding model fine-tuned on training data, increasing performance by 1.16\% in cIoU and 0.81\% in gIoU metrics. This improvement significantly exceeds that of models like LLaVA-1.5 and mPLUG-PaperOWI by a margin of 5\%-10\%. In integrity verification tasks, despite significant model limitations, the influence of attribute information on task performance is more pronounced than changes resulting from switching the base model. These findings indicate that within our framework, accurate attribute information is essential for aligning text with figure modules and effectively performing integrity verification tasks.

\subsubsection{OCR and NER}
\label{section:A}

\begin{table}[h!]
\centering
\caption{Performance variations in the integrity verification task when employing different OCR and NER modules. 'Ideal Model' refers to the performance exhibited upon fully recognizing all modules within figures and terms in the text. SciBert* is a Bert model fine-tuned for text integrity verification.}\label{tbl3}
\begin{tabular}{ccccccc}
\hline
Type &Model & \multicolumn{5}{c}{Integrity Verification}\\
\cline{3-7}
&& \multicolumn{2}{c}{Image}& \multicolumn{3}{c}{Text}\\
\cline{3-7}
& & cIoU $\uparrow$ & gIoU $\uparrow$ & P $\uparrow$ & R $\uparrow$ & F1 $\uparrow$\\
\hline
\multirow{3}*{OCR Module}&mPLUG-PaperOwl & 0.81 & 1.44 & 7.00 & 2.00  & 3.00\\
&Paddle-OCR & 17.54  & 17.35 & 23.0 & 30.00 & 26.00 \\
& Ideal model &  28.15 & 32.92 & 69.00 & 57.00 & 62.00 \\
\hline
\multirow{3}*{NER Module}&SciBert* & 24.17 & 23.78 & 49.00 & 37.00 & 42.00\\
&GPT-4 & 24.17 & 23.78 & 51.00 & 37.00 & 43.00 \\
& Ideal model & 24.17 & 23.78 & 67.00  & 37.00 & 48.00\\

\hline
\end{tabular}
\end{table}

Beyond evaluating the direct impact of our model on integrity verification tasks, we assess the broader influence of OCR (Optical Character Recognition) and NER (Named Entity Recognition) performance, which serve as crucial upstream components in our pipeline. As demonstrated in Table \ref{tbl3}, the performance of the OCR model significantly affects task outcomes, where using GPT-4 as an OCR tool falls short of the desired F1 score by 19\%. The more accurately the OCR model performs, the more figure modules our framework is able to correctly identify. Conversely, the better the NER model performs, the more accurately the framework recognizes modules that are not described in the text. Despite the relative maturity of NER technologies, employing GPT-4 or SciBert trained on text integrity verification data narrows the performance gap to only 5\% from the ideal. This indicates that under the constraints of text-figure alignment model performance, NER model improvements do not markedly influence overall task outcomes.

\subsubsection{Voting mechanism performance}
\begin{table}[h!]
\centering
\caption{Comparison of model performance with the voting mechanism versus simple attribute information integration. The analysis focuses on the model's performance as parameter $\alpha$ varies from 1 to 3.}\label{tbl4}
\scalebox{0.95}{
\begin{tabular}{cccccc}
\hline
Type & Model &\multicolumn{2}{c}{Text-figure Alignment} & \multicolumn{2}{c}{Integrity Verification}\\
\cline{3-6}
& & cIoU $\uparrow$ & gIoU $\uparrow$ & cIoU $\uparrow$ & gIoU $\uparrow$\\
\hline
\multirow{3}*{Integration}& EPM ($\alpha=1, \beta=1$)&65.84 (66.48) &  87.33 (75.75)&24.17 (28.48) & 23.78 (32.22)\\
& EPM ($\alpha=2, \beta=1$) &72.87 (74.73) & 90.07 (82.25)& 22.22 (27.05)  & 26.61 (35.85)\\
 &  EPM ($\alpha=3, \beta=1$) & 67.09 (73.09) & 88.35 (83.50) & 25.86 (28.22) & 27.59 (36.15)\\
\hline
\multirow{3}*{Voting}& EPM ($\alpha=1, \beta=1$)  &67.37 (68.41) & 87.74 (76.48)& 24.67 (26.33) & 23.88 (30.71)\\
& EPM ($\alpha=2, \beta=1$) &73.80 (75.74) & 90.14 (82.18)& 23.02 (26.95) & 26.81 (36.03)\\
 & EPM ($\alpha=3, \beta=1$) & 69.49 (74.99) & 88.41 (83.01) & 23.88 (27.36)  & 27.51 (36.38)\\

\hline
\end{tabular}}
\end{table}
The analysis of the figure interpreter reveals that accurate identification of attributes critically influences the text-figure alignment performance. Erroneous attribute information can mislead the model in aligning text with figures, as shown in the last row of Table \ref{tbl2}. The integration of different types of attribute information—semantic, relative position, and absolute position—enhances the segmentation task's performance by approximately 1\% each. However, when these attributes are combined within the model, interference occurs among the information types, resulting in an overall performance that does not surpass the highest performance achieved by any single attribute type.

Consequently, rather than merging all information directly, we adopt a simplistic voting mechanism. This approach treats each type of information as an independent input for figure segmentation modules and averages their outcomes cumulatively. According to Table \ref{tbl4}, employing this voting mechanism in the module identification task leads to a 1\%-2\% improvement in the cumulative Intersection over Union (cIoU) metric and up to a 0.41\% increase in the generalized Intersection over Union (gIoU) metric, compared to parameter fusion. This strategy uniformly enhances the recognition performance across most figure modules.

Nevertheless, the performance comparison between this voting mechanism and direct fusion reveals distinct strengths and weaknesses in the integrity verification task, primarily reflecting the inherent limitations of the model itself. These findings underscore the need for careful consideration of how different attribute information is integrated into the model to optimize both module identification and integrity verification performance.

\begin{figure}[h!]
\centering
\subfigure[Performance variations induced by the sampling rate $\alpha$.]{
\begin{minipage}[t]{0.45\textwidth}
\centering
% \label{FIG:sample(a)}
\includegraphics[scale=0.25]{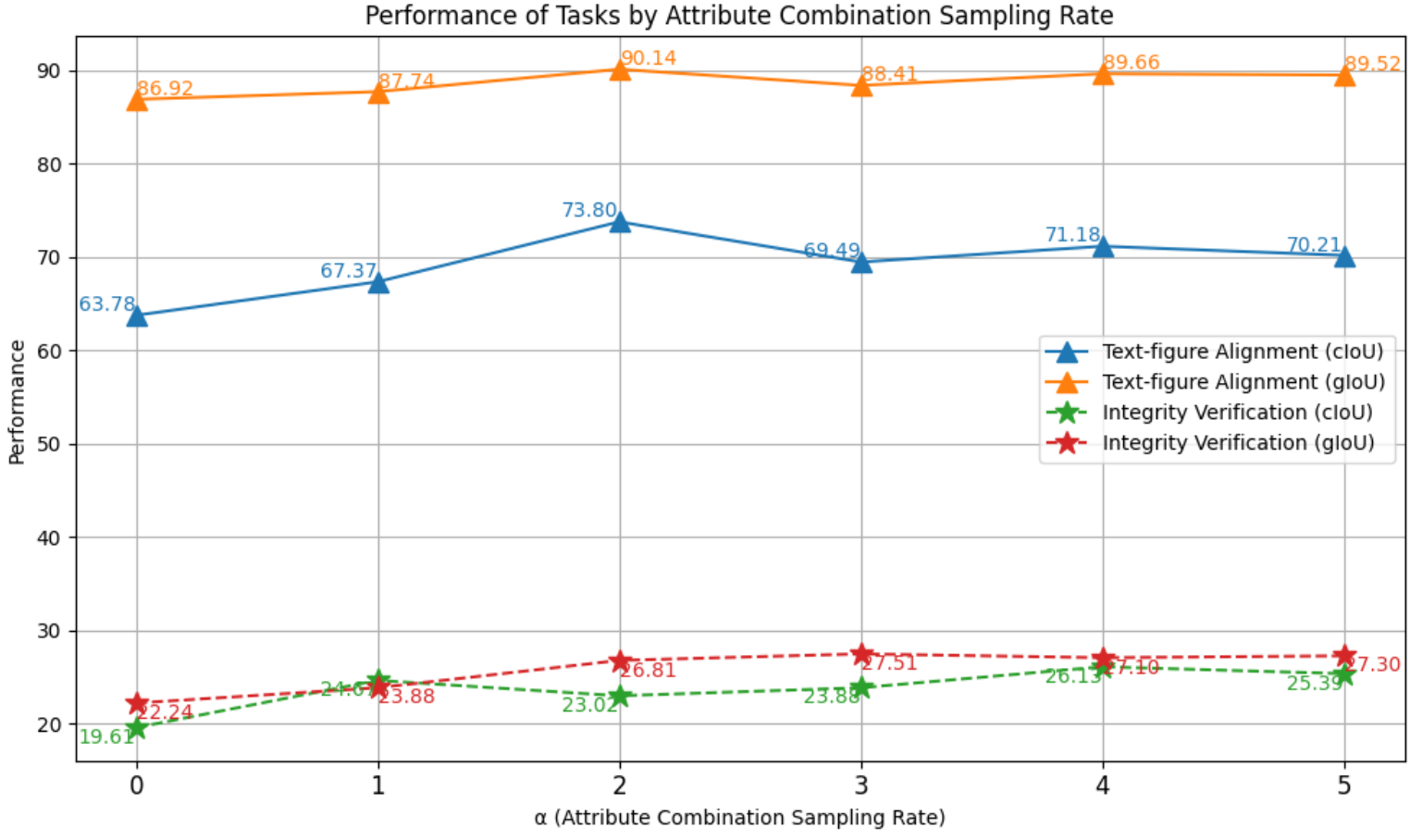}
%\caption{fig1}
\end{minipage}
}
\hfill
\subfigure[Performance variations induced by the sampling rate $\beta$.]{
\begin{minipage}[t]{0.45\textwidth}
\centering
\includegraphics[scale=0.25]{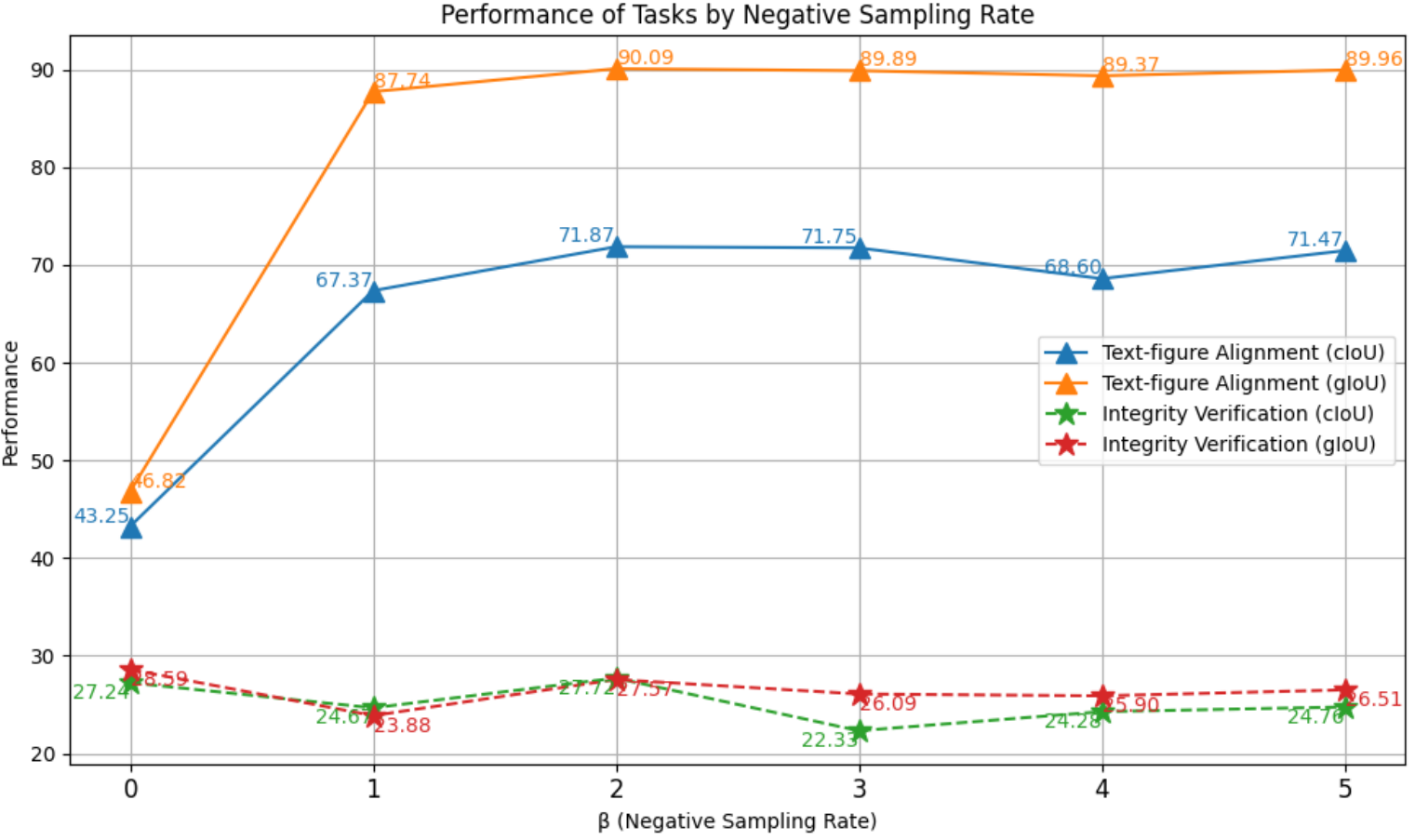}
%\caption{fig2}
\end{minipage}
}
\centering
% \label{FIG:sample(b)}
\caption{Variations in model performance across text-figure alignment and integrity verification tasks with changes in the attribute combination sampling rate $\alpha$ and negative sampling rate $\beta$.}
\label{FIG:sample}
\end{figure}

\subsection{Effectiveness of sample}
\label{section:H}
To optimize the model's ability to comprehend spatial and semantic information and to effectively discriminate against non-existent modules, we implement sampling for both positive and negative samples during the training process. This section investigates the impact of varying the sampling rates, $\alpha$ and $\beta$, which range from 1 to 5. As depicted in Figure \ref{FIG:sample}(a), introducing attribute combination sampling notably improves the model's performance in text-figure alignment and integrity verification tasks. Specifically, we observe improvements of at least 3.5\% and 5\% in the cumulative Intersection over Union (cIoU) metric for these tasks, respectively. The model's performance in text-figure alignment peaks when the sampling rate $\alpha$ is set to 2, while its performance in integrity tasks tends to continuously increase with higher $\alpha$ values. Figure \ref{FIG:sample}(b) highlights that negative sampling significantly enhances the model's ability to discern negative samples. However, using an excessive number of negative samples can detrimentally impact the overall performance, particularly in integrity verification tasks. Interestingly, setting $\beta$ to 2 allows the model's performance on both text-figure alignment and integrity verification tasks to reach optimal levels.

\subsection{Transferability}
\label{section:C}

\begin{table}[h!]
\centering
\caption{Changes in model performance following the injection of attribute information into the baseline model.}\label{Tbl:transfer_model}
\begin{tabular}{cccccc}
\hline
 Dataset & Model & \multicolumn{2}{c}{Text-figure Alignment} & \multicolumn{2}{c}{Integrity Verification}\\
\cline{3-6}
& & cIoU $\uparrow$ & gIoU $\uparrow$ &   cIoU $\uparrow$ & gIoU $\uparrow$ \\
\hline
\multirow{4}{*}{Figure-seg}&GRES & 1.23 \textcolor{red}{(+0.09)} & 4.77 \textcolor{red}{(+1.29)} & 4.04 \textcolor{blue}{(-0.94)}& 1.47 \textcolor{red}{(+0.27)}\\
&X-Decoder & 4.71 \textcolor{red}{(+0.37)} & 30.84 \textcolor{red}{(+0.33)} & 1.57 (-1.68) & 1.04 \textcolor{blue}{(-0.11)} \\
&SEEM & 4.57 \textcolor{red}{(+0.60)} & 12.88 \textcolor{red}{(+0.65)} & 4.62 \textcolor{blue}{(-0.35)} & 3.21 \textcolor{red}{(+0.18)}\\
&FastSAM & 36.67 \textcolor{blue}{(-14.60)} & 49.13 \textcolor{red}{(+4.26)} & 24.09 \textcolor{red}{(+6.46)} & 27.42 \textcolor{red}{(+5.20)}\\
\hline
\end{tabular}
\end{table}

\begin{table}
\centering
\caption{Changes in model performance before and after the injection of attribute information in the task of referring segmentation.}\label{Tbl:transfer_dataset}
\begin{tabular}{cccccc}
\hline
 Dataset & Model & \multicolumn{2}{c}{Referring Segment (Before)} & \multicolumn{2}{c}{Referring Segment (After)}\\
\cline{3-6}
& & cIoU $\uparrow$ & gIoU $\uparrow$ &   cIoU $\uparrow$ & gIoU $\uparrow$ \\
\hline
\multirow{5}{*}{RefCOCOg}&GRES & 9.98 & 5.05 & 10.17 \textcolor{red}{(+0.19)} & 5.18 \textcolor{red}{(+0.13)}\\
&X-Decoder & 65.85 & 70.30& 66.33 \textcolor{red}{(+0.48)}& 70.14 \textcolor{blue}{(-0.16)}\\
&SEEM & 66.04 & 70.19& 65.99 \textcolor{blue}{(-0.05)}& 70.38 \textcolor{red}{(+0.19)}\\
&FastSAM & 18.27 & 24.53& 17.75 \textcolor{blue}{(-0.52)} & 24.07 \textcolor{blue}{(-0.46)}\\
\hline
\end{tabular}
\end{table}

The results discussed previously confirm the effectiveness of our method in understanding scientific figures. In this section, we extend our investigation to the transferability of our approach to other multimodal models and natural images. As shown in Table \ref{Tbl:transfer_model}, the application of attribute injection and a voting mechanism generally enhances the performance of text-figure alignment tasks across various baseline models, with the notable exception of FastSAM. However, results for integrity verification tasks are mixed, with FastSAM showing significant improvements, which underscores its unique responsiveness to attribute information. These outcomes indicate that while attribute information can enhance the figure comprehension of multimodal models to some degree, for models with limited capabilities in this area, the additional attribute information provides minimal benefits and may even introduce confusion.

Table \ref{Tbl:transfer_dataset} displays the performance of these models on the RefCOCOg dataset \cite{mao2016generation}, where texts already include detailed descriptions of image attributes. In this context, we augment the textual descriptions with additional data on color and position using the LLaVA model. The impact of our method on model performance in this dataset is minimal, likely due to the already rich attribute information within the text and potential inaccuracies in the attribute enhancements provided by the image understanding models. Despite these challenges, as the image comprehension capabilities of multimodal models improve, enriching them with more detailed information is anticipated to better the accuracy of text-figure multimodal alignment.

\begin{figure}[h!]
	\centering
		\includegraphics[scale=.85]{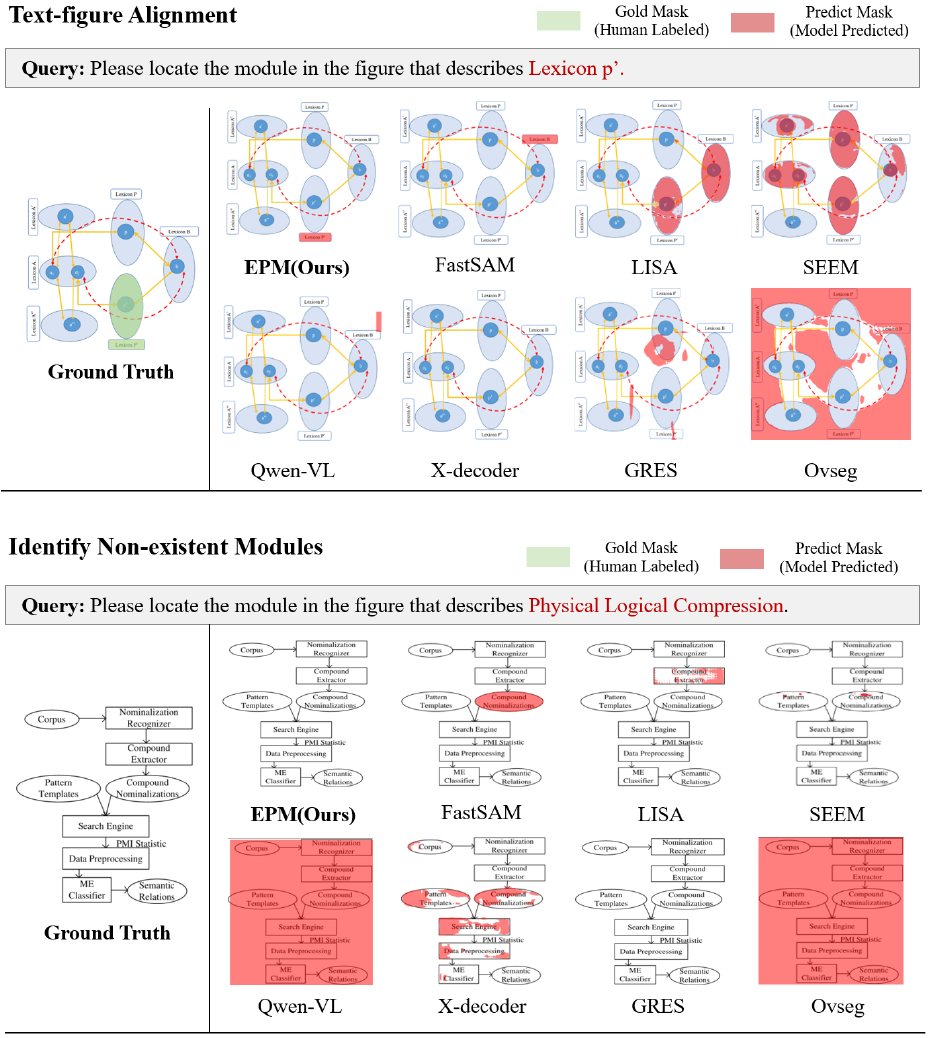}
	\caption{Comparison with the baseline model in terms of figure text-figure alignment capabilities. Case 1 \cite{dranca2020multi} demonstrates the model's ability to distinguish between similar modules, while Case 2 \cite{zhao2007semantic} showcases the model's capability to identify non-existent modules.}
	\label{FIG:case_study_ms}
\end{figure}

\begin{figure}[h!]
	\centering
		\includegraphics[width=0.85\textwidth]{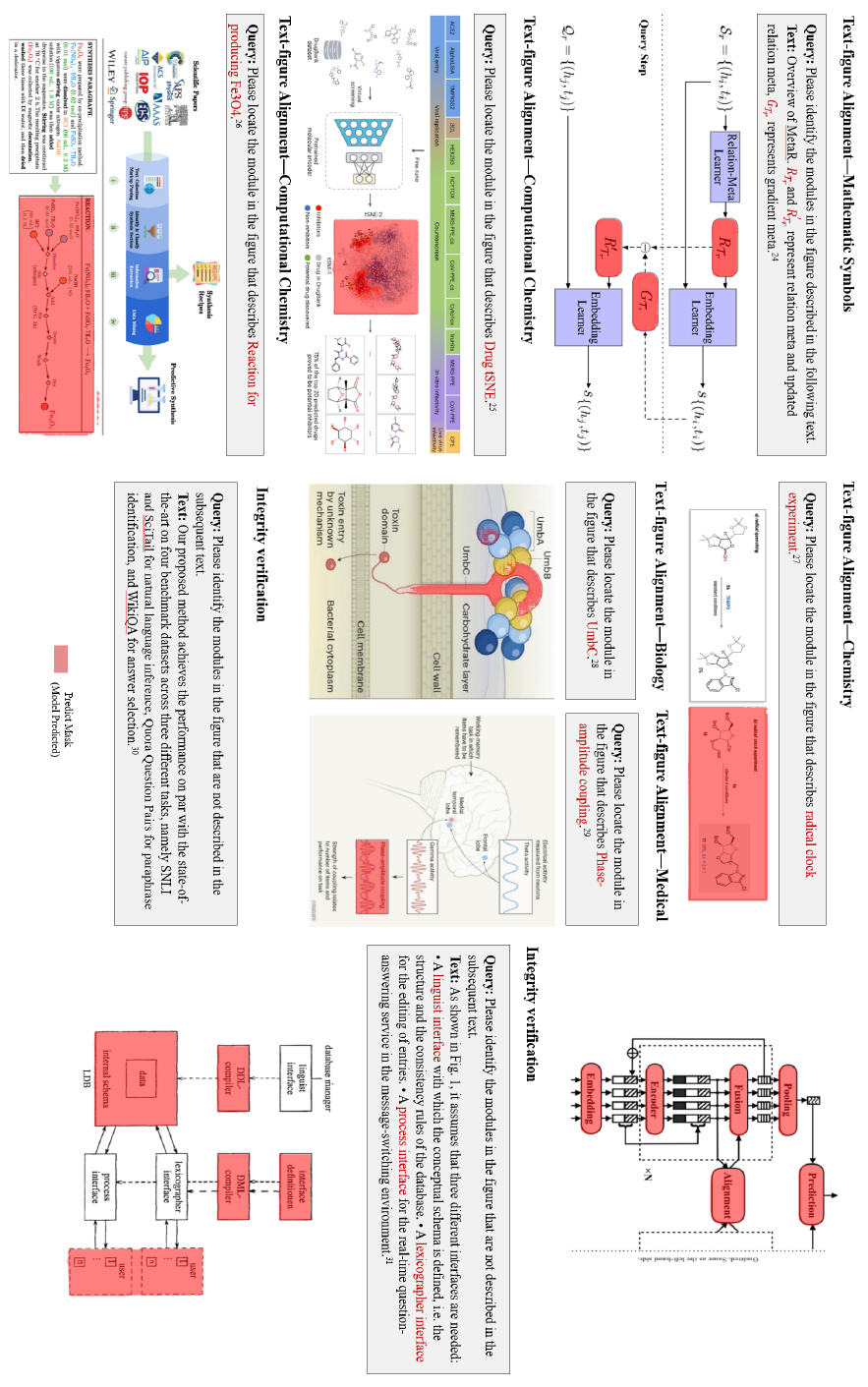}
	\caption{Examples of segmenting and verifying various types and domains of scientific figures. Case 1 \cite{chen2019meta} demonstrates an example of aligning figure modules with text. Cases 2 \cite{zeng2022accurate} and 3 \cite{wang2022dataset} showcase segmentation results of scientific figures from computational chemistry research. Cases 4 \cite{liu2024dehydroxylative}, 5 \cite{coulthurst2024bacteria}, and 6 \cite{williams2024coupled} display segmentation results of chemical formulas and biological structure diagrams in the fields of chemistry, biology, and medical research, respectively. Cases 7 \cite{yang2019simple} and 8 \cite{domenig1986towards} present the results of integrity verification on framework diagrams and flowcharts.}
	\label{FIG:case_study_iv}
\end{figure}

\begin{figure}[h!]
	\centering
		\includegraphics[scale=.85]{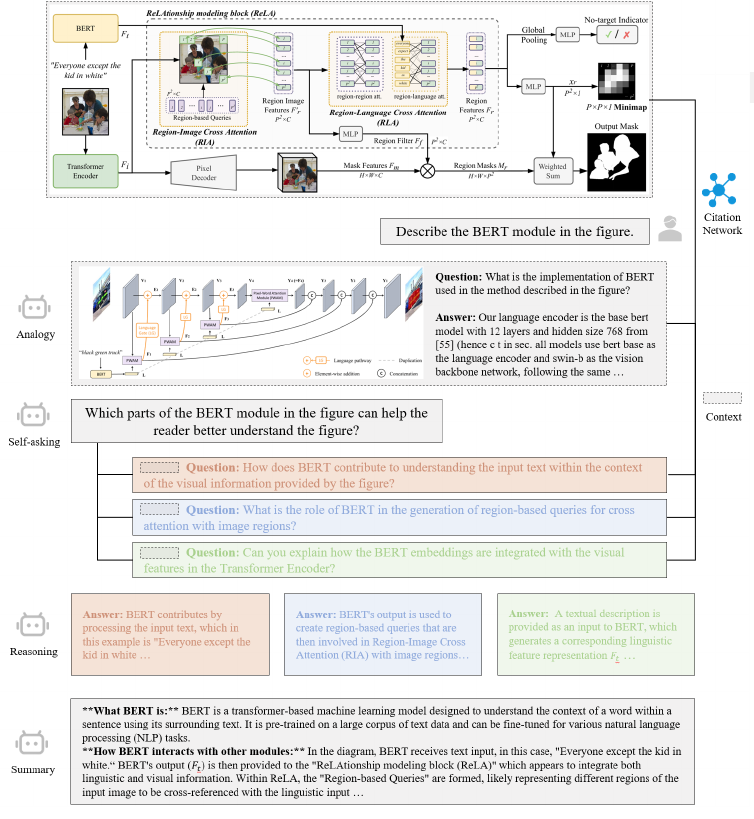}
	\caption{Example of integrity augmentation by supplementing the description of the module in the figure through analogical reasoning. The figures in the example are from \textbf{GRES} \cite{liu2023gres}, and the analogous figures are from the citations of it \cite{yang2022lavt}.}
	\label{FIG:case_study_ia}
\end{figure}

\subsection{Case study}
\label{section:B}
This section demonstrates our framework's efficacy in performing text-figure alignment and integrity verification across various types of scientific figures. As illustrated in Figure \ref{FIG:case_study_ms}, we present a comparison between our framework and other methodologies in aligning and detecting non-existent visual elements. It is evident that our model can accurately locate modules differing by only one or two characters and can precisely identify elements that are absent in the figures. Figure \ref{FIG:case_study_iv} showcases, from left to right, the effectiveness of our framework in segmenting entities containing mathematical symbols, chemical formulas, and biological units. Additionally, it is capable of effectively identifying unexplained sections in flowcharts through joint analysis of text and images. As shown in Figure \ref{FIG:case_study_iv}, we display an example of utilizing a citation network to construct an analogical reasoning process that guides MLLMs to enhance the completeness of figures.

While our method achieves strong performance in understanding complex scientific figures, it is not without limitations. For instance, it struggles with accurately identifying multiple modules within a figure simultaneously and occasionally falsely detects modules already mentioned in the text.

\section{Discussion}

\subsection{Theoretic implications}
This study delves into the research problems of assisting the reading and writing of scientific papers, under the broader research framework of integrity verification proposed in prior work \cite{shi2024integrity}. Specifically, it designs a targeted integrity verification task and methodological strategies for scientific diagrams rich in semantic information. Unlike existing research on interpreting natural images and data charts, this study pioneers in focusing on more complex process diagrams and framework diagrams, constructing a model framework for fine-grained alignment between paper texts and diagrammatic modules based on multi-dimensional information such as the position and function of modules within diagrams. Furthermore, the study goes beyond simply how scientific diagrams are referenced in the text, to how unrecognized modules can be identified, aiding researchers in better understanding and describing scientific diagrams. This research enriches the overarching theoretical framework of integrity verification, establishing new theoretical foundations and pathways for understanding complex scientific diagrams.

\subsection{Practical implications}

This study introduces a MLLM-based method for the segmentation and integrity verification of modules within scientific diagrams. Extensive empirical evidence demonstrates the effectiveness of our approach in understanding complex scientific diagrams such as flowcharts and framework diagrams, showcasing good scalability and domain transferability. This offers new insights into the development of MLLMs for the interpretation of scientific diagrams. Secondly, we design a specialized dataset, \textit{Figure-seg}, specifically for the task of scientific diagram text-figure alignment and integrity verification, thereby establishing a foundation for subsequent research and performance enhancement in this area. Moreover, building upon the methodological research, we have designed an application to assist researchers in comprehending scientific diagrams, enabling direct links between diagrammatic module units and the textual content of papers. Based on this, the study extends into various application research areas such as paper diagram quality assessment and automatic scientific diagram construction, showing significant application value in areas like paper writing assistance and diagram generation.

\subsection{Limitation}
Despite our method achieving better results than existing models in the tasks of scientific diagram segmentation and integrity verification, there remain certain limitations as described below.

\begin{itemize}
    \item \textbf{Segment anything within figures.} Identifying all meaningful modules within a figure is foundational to discovering undiscussed modules. Currently, this has not been effectively achieved in all SAM series models, such as FastSAM and EfficientSAM, where a significant number of meaningless or falsely detected modules considerably impact the overall method performance. 
    
    \item \textbf{Alignment of text and processes in diagrams.} Currently, our method preliminarily achieves a fine-grained understanding of diagrams through the alignment of terms in the text with modules in the diagrams. However, merely identifying modules within the diagrams does not adequately restore the semantic information of the diagrams, especially for flowcharts, where the semantic relationships between modules are particularly crucial. This aspect has not been effectively addressed in our research.
\end{itemize}

\subsection{Future work}
In the future, we will expand upon the existing research from both theoretical and methodological perspectives.

\begin{itemize}
    \item \textbf{Theoretical work.} We will further enrich and expand the integrity verification framework, embarking from multiple dimensions such as paper structure, textual content, and diagrammatic content, to conduct fine-grained interpretations of scientific papers. This will establish a new theoretical foundation for the analysis and bibliometrics of scientific literature.
    
    \item \textbf{Methodological work.} We aim to improve the performance and execution efficiency of the model, addressing the current methods' inadequacies in detecting unexplained modules. Furthermore, we plan to establish an end-to-end MLLM for the understanding and interpretation of complex scientific diagrams.
\end{itemize}
\section{Conclusion}
In this study, within the theoretical framework of integrity verification, we introduce the task of integrity verification specifically designed for scientific diagrams in scientific papers. Around this task, we construct a specialized dataset, Figure-seg, and develop an MLLM-based framework, EPM, to perform the task. Extensive experiments demonstrate that our proposed approach outperforms the current SOTA models in understanding and detecting the integrity of complex scientific diagrams such as flowcharts and framework diagrams, with performance improvements in text-figure alignment and undepicted module detection improved by 22.53\% and 4.90\% in cIoU metric, and 45.13\% and 4.52\% in gIoU metric, respectively. 
Furthermore, based on the results of integrity verification, we design a method for completing the integrity of scientific figures inspired by analogical reasoning, providing new theoretical and methodological foundations for research in this field.

\bibliographystyle{unsrt}

\bibliography{cas-refs}

\appendix
\section{Appendix}

\subsection{Data distribution}
\label{section:G}
\begin{figure}[h!]
	\centering
		\includegraphics[width=0.8\textwidth]{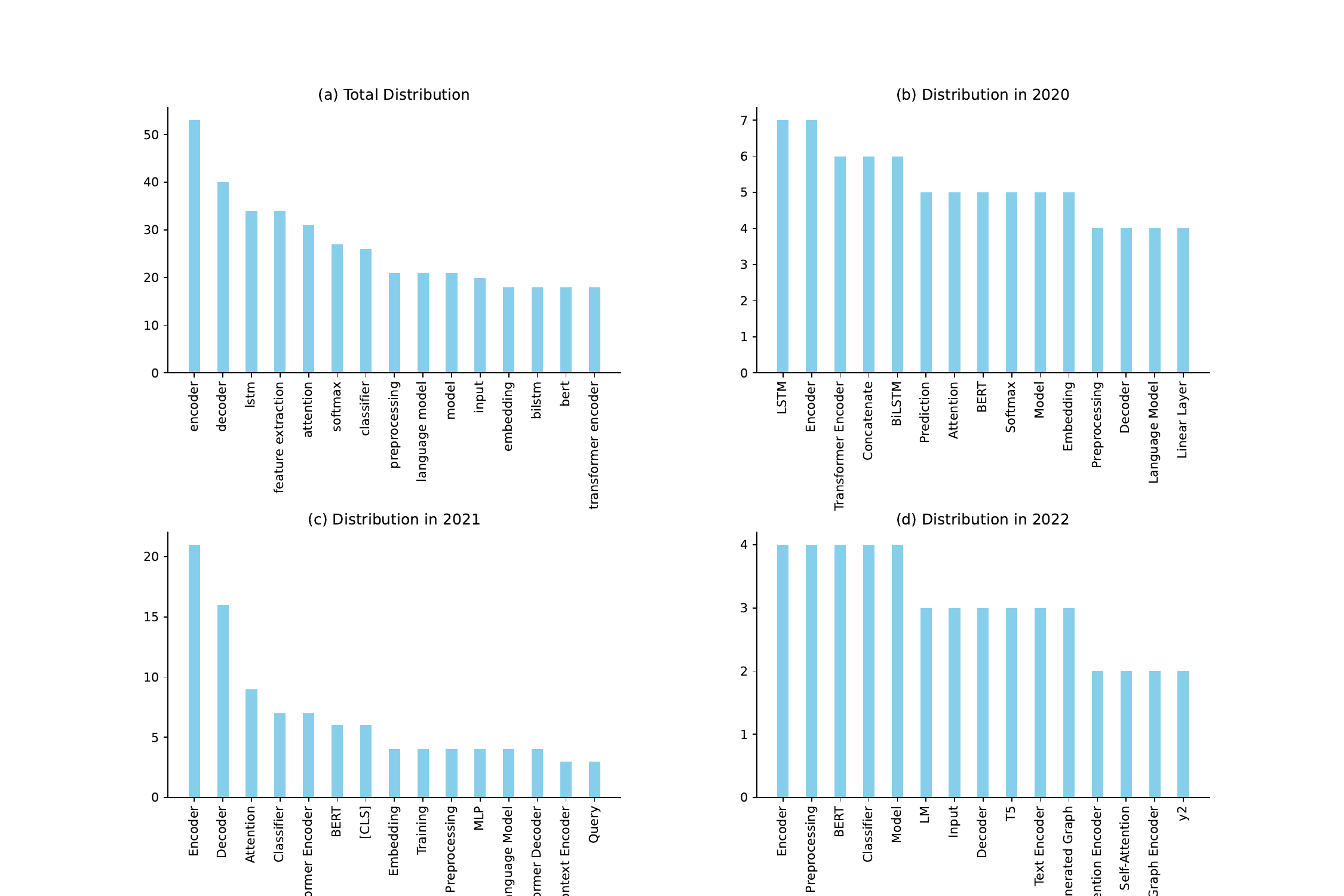}
	\caption{The dataset's top 10 most frequently occurring modules are listed in order of decreasing frequency. Figure (a) displays the overall distribution, while Figures (b), (c), and (d) specifically depict the distributions for the years 2020, 2021, and 2022, respectively.}
	\label{FIG:frequency}
\end{figure}

Building upon conventional data statistics, we further investigate the distribution of annotated module contents within the overall dataset. In Figures \ref{FIG:frequency}, we present the overall frequency of occurrences of modules in scientific diagrams and the high-frequency modules from 2020 to 2022. It is observable that terms such as “encoder”, “decoder”, and “lstm” rank among the top 3 in the frequency of occurrence within the scientific diagrams of ACL papers. This indicates that encoder-decoder architecture are commonly used in the field of natural language processing, employing LSTM (Long Short-Term Memory) networks for feature extraction tasks. Moreover, observing the shift in module popularity over three years, language models like BERT and T5 have garnered more attention from researchers compared to the once-popular LSTM in 2020. Although our dataset does not cover every module in the diagrams, preventing a comprehensive reflection of each term's true frequency, it still suggests, to a certain extent, that scientific diagrams can serve as a key source for understanding domain development and knowledge inheritance beyond textual content. This has significant implications for knowledge management and informetrics research.

% \begin{figure}
% 	\centering
% 		\includegraphics[scale=.5]{figs/time.pdf}
% 	\caption{A word cloud constructed from the frequency of module occurrences in the dataset during the years 2020-2022.}
% 	\label{FIG:A2}
% \end{figure}

\subsection{Parameter setting}
\subsubsection{Data construction}
\label{section:E}
In the construction of the \textit{Figure-seg} dataset, the method for maintaining consistency between FastSAM's generated segmentation masks using text and coordinates is to ensure that the IoU of the results generated by both exceeds 0.95. Besides, the minimum pixel distance $Dist_{pixel}$ for merging two bounding boxes is set to 50. The minimum threshold for the IoU between text boxes and segmentation modules, $min_{iou}$, is set at 0.1.

\subsubsection{Training}
\label{section:F}
\begin{table}[h!]
\centering
\caption{Detailed parameter settings of the model in training phrase.}\label{Tbl:parameter}
\begin{tabular}{ccc}
\hline
Type & Parameter & Value  \\
\hline
 \multirow{3}{*}{Model} & Base model & llava-1.5-13b \\
 & Image encoder & clip-vit-large-patch14 \\
 & Mask decoder & sam\_vit\_h  \\
 \hline
  \multirow{3}{*}{Training} & Epoch $E$& 10 \\
 & Batch size $B$ & 8 \\
 & Learning rate $lr$ & 0.0003  \\
 & Lora\_alpha $l_\alpha$ & 16  \\
 & Lora\_dropout $ld$& 0.95 \\
 & Shift\_CE loss weight $\lambda_1$ & 1.0 \\
 & BCE loss weight $\lambda_2$ & 2.0 \\
 & DICE loss weight $\lambda_3$ & 0.5 \\
 
 \hline
 \multirow{2}{*}{Data} & Attribute sample $\alpha$ & 2 \\
 & Negative sample $\beta$ & 1 \\
\hline
\end{tabular}
\end{table}
The parameters used in the model training process are as shown in Table \ref{Tbl:parameter}: (1) Model configuration: llava-1.5-13b is employed as the base model, clip-vit-large-patch14 serves as the image encoder, and the mask decoder is initialized using the sam\_vit\_h model. (2) Training parameter settings: The epoch is set to 10, the batch size is set to 8, and the learning rate is initialized at 0.0003. (3) Data configuration: The sampling rate for attribute combination is 2. The sampling rate for negative samples is 1 (meaning that for every positive example, one negative example is generated). (4) Adapter settings: lora\_alpha is set to 16, and lora\_dropout is set to 0.95. (5) Loss configuration: The weight for Shift\_CE loss is 1.0, for DICE loss is 0.5, and for BCE loss is 2.0.

\subsection{Instruction}
\subsubsection{Data collection template}
The following is the instruction template used during our dataset construction process:
\begin{tcolorbox}[colback=gray!10, boxrule=0pt, title=Data collection template, colbacktitle=gray!40]
\textbf{SYSTEM:} You are a very helpful language and visual assistant. You can understand the visual content in scientific figures within the scientific literature, aiding users in correlating descriptions from the scientific literature with the modules in the figures. \\

\textbf{QUERY (SEMANTIC):} \textit{[image]} Please describe the function of the module '\%s' in the diagram in one sentence using the format: Its function is XX. If you cannot identify the module from the picture, please directly answer 'Unknown'. \\

\textbf{QUERY (SPATIAL):} \textit{[image]} Please tell me the position of the module '\%s' in the figure using the following format: Its absolute position is: XX, and its relative position is: XX. If you cannot identify the module from the picture, please directly answer 'Unknown'. 

\end{tcolorbox}

\subsubsection{Training template}
The following is the instruction template used during our model training process:
\begin{tcolorbox}[colback=gray!10, boxrule=0pt, title=Training template, colbacktitle=gray!40]
\textbf{SYSTEM:} You are a very helpful language and visual assistant. You can understand the visual content in scientific figures within the scientific literature, aiding users in correlating descriptions from the scientific literature with the modules in the figures. \\

\textbf{QUERY:} \textit{[image]} Segment the corresponding module from the figure based on the given attributes: name: [NAME], function: [FUNCTION], relative position: [RELATIVE POSITION], absolute position: [ABSOLUTE POSITION]. \\

\textbf{ANSWER (POSITIVE):} [MODULE] is the module that has been segmented.\\ 

\textbf{ANSWER (NEGATIVE):} [MODULE] is the module that has been segmented. There is no corresponding module in the figure.

\end{tcolorbox}

\subsubsection{Inference template}
In the inference process, our method utilizes the same instruction template as used during the training phase. For baseline models, to fully leverage their capabilities in understanding figures, we perform multiple rewrites of the input and record the performance that yields the best results. Taking the LISA model as an example:
\begin{tcolorbox}[colback=gray!10, boxrule=0pt, title=Inference template, colbacktitle=gray!40]

\textbf{QUERY:} \textit{[image]} Can you segment the [NAME] in this image? \\

\textbf{METRICS:} cIoU: 27.39 gIou: 41.57  \textcolor{red}{$\times$}\\

\textbf{QUERY:} \textit{[image]} Can you segment the module named [NAME] in this image? \\

\textbf{METRICS:} cIoU: 36.72 gIou: 48.24  \textcolor{green}{$\checkmark$}\\

\textbf{QUERY:} \textit{[image]} Can you segment the [NAME] module in this image? \\

\textbf{METRICS:} cIoU: 36.17 gIou: 46.51  \textcolor{red}{$\times$}\\

\end{tcolorbox}

\begin{figure}[ht!]
    
    \begin{minipage}{0.65\textwidth}
        \begin{algorithm}[H]
    \caption{Simplified Version of Chain-of-Attribute (CoA)}
    \begin{algorithmic}[1]
    \STATE \textbf{Input:} $D^{+}$: Positive examples from segmentation data, $D^{-}$: Negative examples sampled from segmentation data, $G$: Figure, $M$: Module name, $\mathcal{I}$: Figure interpreter trained on $D^{+}$ and $D^{-}$
    \STATE \textbf{Output:} $\textit{Mask}_{final}$: Final segmentation
    
    \STATE $(attr_{abs}, attr_{rel}, attr_{sem}) \gets \mathcal{I}(G, M)$
        \STATE \ \ \ \textcolor{gray}{// Output module $M$'s attributes using interpreter $\mathcal{I}$}
        \STATE $\textit{Mask}_{abs} \gets \mathcal{M}(G, \textit{Concat}(M, attr_{abs}))$ 
        \STATE $\textit{Mask}_{rel} \gets \mathcal{M}(G, \textit{Concat}(M, attr_{rel}))$ 
        \STATE $\textit{Mask}_{sem} \gets \mathcal{M}(G, \textit{Concat}(M, attr_{sem}))$ 
        \STATE \ \ \ \textcolor{gray}{// Obtain segmentations by feeding attributes to $\mathcal{M}$}
        \STATE $\textit{Mask}_{final} \gets \text{Vote}(\textit{Mask}_{abs}, \textit{Mask}_{rel}, \textit{Mask}_{sem})$ 
        \STATE \ \ \ \textcolor{gray}{// Synthesize final mask by voting among $\textit{Masks}$}
    \end{algorithmic}
\end{algorithm}
    \end{minipage}
    \hfill
    \subfigure[Simplified process of CoA]{
    \begin{minipage}{0.3\textwidth}
        \centering
        \includegraphics[width=\textwidth]{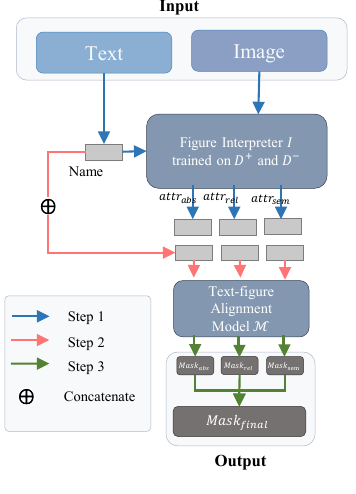} % Replace 'example-image-a' with your image file name
        
    \end{minipage}}
    \caption[width=0.5\textwidth]{Detailed explanation of the simplified Chain-of-Attribute (CoA) reasoning process.}
    \label{FIG:simplified_coa}
\end{figure}

% \begin{algorithm}
%     \caption{Simplified Version of Chain-of-Attribute (CoA)}
%     \begin{algorithmic}[1]
%     \STATE \textbf{Input:} $D^{+}$: Positive examples from segmentation data, $D^{-}$: Negative examples sampled from segmentation data, $G$: Figure, $M$: Module name, $\mathcal{I}$: Figure interpreter trained on $D^{+}$ and $D^{-}$
%     \STATE \textbf{Output:} $\textit{Mask}_{final}$: Final segmentation
    
%     \STATE $(attr_{abs}, attr_{rel}, attr_{func}) \gets \mathcal{I}(G, M)$
%         \STATE \ \ \ \textcolor{gray}{// Output module $M$'s attributes using interpreter $\mathcal{I}$}
%         \STATE $\textit{Mask}_{abs} \gets \mathcal{M}(G, \textit{Concat}(M, attr_{abs}))$ 
%         \STATE $\textit{Mask}_{rel} \gets \mathcal{M}(G, \textit{Concat}(M, attr_{rel}))$ 
%         \STATE $\textit{Mask}_{func} \gets \mathcal{M}(G, \textit{Concat}(M, attr_{func}))$ 
%         \STATE \ \ \ \textcolor{gray}{// Obtain segmentations by feeding attributes to $\mathcal{M}$}
%         \STATE $\textit{Mask}_{final} \gets \text{Vote}(\textit{Mask}_{abs}, \textit{Mask}_{rel}, \textit{Mask}_{func})$ 
%         \STATE \ \ \ \textcolor{gray}{// Synthesize final mask by voting among $\textit{Masks}$}
%     \end{algorithmic}
% \end{algorithm}

\subsection{The design philosophy of the CoA}
\begin{table}[h!]
\centering
\caption{Comparison of CoA and Simplified CoA in terms of performance and inference efficiency. We report the average time taken to segment each module in the text-figure alignment task and the average time to check each pair of text and figure in the integrity verification task. The testing of inference efficiency is conducted on a single A100 80G GPU.}\label{Tbl:pipeline}
\begin{tabular}{ccccccc}
\hline
Model & \multicolumn{3}{c}{Text-figure Alignment} & \multicolumn{3}{c}{Integrity Verification}\\
\cline{2-7}
 & cIoU $\uparrow$ & gIoU $\uparrow$ & time$\downarrow$ &   cIoU $\uparrow$ & gIoU $\uparrow$ & time$\downarrow$ \\
\hline
CoA & 67.37 & 87.74 & 3.68s/module & 24.67 & 23.88 & 132.87s/entry\\
Simplified CoA & 65.52 & 85.64 & 5.30s/module & 22.92 & 23.95 &97.63s/entry\\
\hline
\end{tabular}
\end{table}

In the CoA reasoning process proposed in this study, the text-figure alignment model $\mathcal{M}$ is invoked twice for detecting the existence of modules and inferring their Masks. This step, as depicted in Figure \ref{FIG:simplified_coa}, is indeed amenable to simplification. The underlying concept involves partially transferring the responsibility of determining module existence to the figure interpreter, which assesses the presence of corresponding modules and attributes within the figure. Following this assessment, theline segmentation task is delegated to model $\mathcal{M}$. However, as shown in Table \ref{Tbl:pipeline}, while this approach accelerates the inference process (approximately 35 seconds faster per data entry, CoA's speed advantage in text-figure alignment stems from faster negative sample detection), the introduction of negative examples during the training of the figure interpreter enhances image discrimination capabilities at the expense of attribute recognition ability, leading to overall performance that is inferior to the initial strategy employed in our research. This further corroborates that the image understanding and segmentation modules constitute the most pivotal components of our entire pipeline, with their capabilities being decisive for the overall performance and operational efficiency.

\end{document}